\documentclass[12pt,authoryear]{myelsarticle}

\usepackage{graphicx}
\usepackage{amssymb}
\usepackage{amsbsy} 
\usepackage{lineno}
\usepackage{hyperref}
\usepackage{multirow}
\usepackage[table]{xcolor}
\usepackage[usenames,dvipsnames]{pstricks}
\usepackage{fancyhdr}

\journal{ISPRS J. Photo. Remote Sens.}

\begin{document}

\begin{frontmatter}




\title{Land cover mapping at very high resolution with rotation equivariant CNNs: towards small yet accurate models }
\author[1,2]{Diego Marcos}\cortext[]{Corresponding Author: Diego Marcos, diego.marcos@wur.nl}
\author[1]{Michele Volpi}
\author[2]{Benjamin Kellenberger}
\author[1,2]{Devis Tuia}

\address[1]{MultiModal Remote Sensing, University of Zurich, Switzerland.\\\href{http://www.geo.uzh.ch/en/units/multimodal-remote-sensing}{www.geo.uzh.ch/en/units/multimodal-remote-sensing}}
\address[2]{Laboratory of GeoInformation Science and Remote Sensing, Wageningen University and Research, the Netherlands. \href{www.geo-informatie.nl}{www.geo-informatie.nl}}

\begin{abstract}
\textbf{Work published on ISPRS Journal of Photogrammetry and Remote Sensing, DOI: 10.1016/j.isprsjprs.2018.01.021}

In remote sensing images, the absolute orientation of objects is arbitrary. Depending on an object's orientation and on a sensor's flight path, objects of the same semantic class can be observed in different orientations in the same image. Equivariance to rotation, in this context understood as responding with a rotated semantic label map when subject to a rotation of the input image, is therefore a very desirable feature, in particular for high capacity models, such as Convolutional Neural Networks (CNNs). If rotation equivariance is encoded in the network, the model is confronted with a simpler task and does not need to learn specific (and redundant) weights to address rotated versions of the same object class. In this work we propose a CNN architecture called Rotation Equivariant Vector Field Network (RotEqNet) to encode rotation equivariance in the network itself. By using rotating convolutions as building blocks and passing only the  the values corresponding to the maximally activating orientation throughout the network in the form of orientation encoding vector fields, RotEqNet treats rotated versions of the same object with the same filter bank and therefore achieves state-of-the-art performances even when using very small architectures trained from scratch. We test RotEqNet in two challenging sub-decimeter resolution semantic labeling problems, and show that we can perform better than a standard CNN while requiring one order of magnitude less parameters.
\end{abstract}

\begin{keyword}
Semantic labeling \sep Deep learning \sep Rotation invariance \sep Sub-decimeter resolution 
\end{keyword}

\end{frontmatter}


\section{Introduction}
\label{S:1}


In this paper we consider the task of \emph{semantic labeling}, which corresponds to the automatic assignment of each pixel to a set of predefined land-cover or land-use classes. The classes are selected specifically for the task to be solved and define the learning problem for the model. 

When using low- to mid-resolution multispectral imagery (e.g. Landsat), it is customary to assume that the spectral information carried by a pixel is sufficient to classify it into one of the semantic classes, thus reducing the need for modeling spatial dependencies. However, when dealing with very-high spatial resolution (VHR) imagery, i.e. imagery in the meter to sub-decimeter resolution range, the sensor trades off spectral resolution to gain spatial details. Such data is commonly composed of red-green-blue (RGB) color channels, occasionally with an extra near infrared (NIR) band. Due to this trade-off, single pixels tend not to contain sufficient information to be assigned with high confidence to the correct semantic class, when relying on spectral characteristics only. Moreover, depending on the task, some classes can be semantically ambiguous:  a typical example is land use mapping, where objects belonging to different classes can be composed of the same material (e.g. road and parking lots), thus making analysis based on spectra of single pixels not suitable. To resolve both problems, spatial context needs to be taken into account, for example via the extraction and use of textural~\citep{regniers2016supervised}, morphological~ \citep{Mur10,Tui15}, tree-based~ \citep{gueguen2015large} or other types~\citep{Mal14} of spatial features. These features consider the neighborhood around a pixel as part of its own characteristics, and allow to place spectral signatures in context and solve ambiguities at the pixel level~\citep{fauvel2013classifhyper}. 
The diverse and extensive pool of possible features led to a surge in works focusing on the automatic generation and selection of discriminant features~\citep{Har02,Glo05,Tui15}, aimed at preventing to compute and store features that are redundant or not suited for a particular task.  

Another common approach to reduce the computational burden while enforcing spatial reasoning is to extract local features from a support defined by unsupervised segmentation. Also, spatial rules can be encoded by Markov random fields, where spatial consistency is usually enforced by minimizing a neighborhood-aware energy function~\citep{Mos13} or specific spatial relationships between the classes~\citep{Volpi2015b}.

In the situations described above, a successful solution comes at the cost of having to manually engineer a high-dimensional set of features potentially covering all the local variations of the data in order to encode robust and discriminative information. In this setting, there is no guarantee that the features employed are optimal for a given semantic labeling problem. 
These problems raised the interest of the community in solutions avoiding to manually engineer the feature space, solutions that are extensively studied 
under the \emph{deep learning} paradigm. The aim of deep learning is to train a parametric system learning feature extraction \emph{jointly} with a classifier~\citep{goodfellow2016deepbook}, in an end-to-end manner. When focusing on image data, Convolutional Neural Networks (CNNs,~\citet{lecun1998pieee}) are state-of-the-art. Their recognized success follows from new ground-breaking results in many computer vision problems. CNNs stand out thanks to their ability to learn complex problem-specific features, while jointly optimizing a loss (e.g. a classifier, a regressor, etc.). 
Thanks to recent hardware advances accelerating CNN training consistently, as well as the existence of pre-trained models to get started, CNNs have become one of the most studied models in recent remote sensing research dealing with VHR imagery, as we briefly review below. 

The first models proposed studied the effectiveness of translating computer vision architectures 
directly to aerial data for tile classification. In that sense, a single label was retrieved per image tile, thus tackling what in computer vision is called the \emph{image classification problem}\footnote{This is not to be confused with the \emph{semantic labeling} problem we address in this paper, which is the task of attributing a label to every pixel in the tile.}: authors in~\citet{castelluccio2015arxiv} and~\citet{penatti2015cvprw} studied the effect of fine-tuning models trained on natural image classification problems, in order to adapt them quickly to above-head image classification. Their results suggested that such a strategy is relevant for image classification and can be used to reuse models trained on a different modality. Transposing these model in the semantic labeling problem is also possible, typically applying the models using a sliding window centered at each location of the image, as tested in~\citet{DFCA}. However, the authors also came to three important conclusions: \emph{i)} models trained from scratch (in opposition to fine-tuned models from vision) tend to provide better results on specific labeling tasks;  \emph{ii)} by predicting a single label per patch, the one corresponding to the pixel on which the patch is centered, these models are not able to encode explicit label dependencies in the output space and \emph{iii)} the computational overhead of the sliding window approach is extremely large.
Such conclusions support the use of network architectures that have been developed specifically for semantic labeling problems: recent efforts tend to consider \emph{fully convolutional} approaches~\citep{long2015cvpr}, where the CNN does not only predict a single label per patch, but actually provides directly the label map for all the pixels that compose the input tile. The approaches proposed vary from spatial interpolation~\citep{maggiori2017convolutional}, fully convolutional models~\citep{Aud16}, deconvolutions~\citep{volpi2017dense}, stacking activations~\citep{Mag16} to hybridization with other classifiers~\citep{liu2017dense}, but they all are consistent in one observation: fully convolutional architectures drastically reduce the inference time and naturally encode some aspect of output dependencies, in particular learning dependent filters at different scales, thus reducing the need of cumbersome postprocessing of the prediction map.

While these works open endless opportunities for remote sensing image processing with CNNs, they also showed one of the biggest downsides of these models: CNNs tend to need large amounts of ground truth to be trained, and setting up the architecture, as well as selecting hyperparameters, can be troublesome, since cross-validation is often prohibitive in terms of processing time. Note that it is often that case when the number of parameters is larger than the number of training samples, which makes regularization techniques and data augmentation a must-do, at the cost of significantly slowing model training. Our contribution aims at addressing this drawback of CNNs, \emph{i.e.} the large model sizes and need for labels when there is a limited availability of ground truth. 
In this paper, we propose to tackle the problem by exploiting a property of objects and features in remote sensing images: \emph{their orientation is arbitrary}.

Overhead imagery differs from natural images in that the \emph{absolute} orientation of objects and features within the images tends to be irrelevant for most tasks, including semantic labeling. This is because the orientation of the camera in nadir-looking imagery is most often arbitrary. As a consequence, the label assigned to an element in the image should not change if the image is taken with a different camera orientation. We call this property \emph{equivariance}, and it is a property that recently attracted a lot of interest in image analysis~\citep{lei2012rotation,cheng2016learning}. 

Given a rotation operator, $g_\alpha(\cdot)$, we say that a function $f(\cdot)$ is equivariant to rotations if $f(g_\alpha(\cdot)) = g_\alpha(f(\cdot))$, invariant to rotations if $f(g_\alpha(\cdot)) = f(\cdot)$ and, more generally, covariant to rotations if $f(g_\alpha(\cdot)) = h(f(\cdot))$, with $h(\cdot)$ being some function other than $g_\alpha(\cdot)$. Note that, in the case of semantic labeling, the property we are interested in is equivariance, although it becomes invariance if we consider a single pixel at a time. We will therefore use the terms equivariance and invariance interchangeably in this paper.

With CNNs, equivariance to the rotation of inputs can be  approximated by randomly rotating the input images during training, a technique known as \emph{data augmentation} or \emph{jittering}~\citep{leen1995neucomp}. 
If the CNN has enough capacity and has seen the training samples in sufficient number of orientations, it will learn to be invariant to rotations~\citep{lenc2015understanding}.  While this kind of data augmentation greatly increases the generalization accuracy, it does not offer any advantage in terms of model compactness, since similar filters, but with different orientations, need to be learned independently. A different approach, hard coding such invariances within the model, has the two main beneficial effects: first, the model becomes robust to variations which are not discriminative, as a standard CNN with enough filters would learn; and second, model-based invariance can be interpreted as some form of regularization \citep{leen1995neucomp}. This added robustness ultimately lead to models which have high capacity (as high as a standard CNN) but with lower sample complexity.

There has been a recent surge in works that explore ways of encoding model-based rotation invariance in CNNs. \cite{laptev2016ti} perform a rotation of the input image in order to reduce the sample complexity of the problem and \cite{jaderberg2015spatial} extend this to affine transformations. These approaches provide invariance to a global rotation of the input image and not to local relative rotations, and are therefore not very well suited for segmentation tasks. \citet{cohen2016group} encode equivariance to shifts and to rotations by multiples of $90^o$ by tying filter weights, while \citet{zhou2017oriented} use linearly interpolated filters. These two methods are in principle suited for segmentation tasks. The former is limited to invariance to $90^o$ rotations and the latter, although offering more flexibility, has the drawback of requiring a trade-off between the number of rotations and the memory requirements, bringing the authors to use $8$ orientations, at multiples of $45^o$. \citet{worrall2016harmonic} reduce the space of possible filters to combinations of complex harmonic wavelets, thusachieving perfect equivariance to rotations. By doing so, they obtain a more compact internal representation by encoding oriented activations as complex valued feature maps, but at the cost of reducing the expressiveness of each filter.

In this paper, we consider a solution that combines the advantages of these two last methods. Our model applies an arbitrary number of rotated instances of each filter at every location, in a way that each filter activation is composed by \emph{a vector} of activations (as opposed to a scalar in standard CNN), thus representing the activation of each rotated filter. We then propose to max-pool these activations, compressing the information in a simple 2D vector that represents the magnitude and orientation of the maximally activating filter. This allows us to encode  fine-grained rotation invariance (\emph{i.e.} very small angles) and, at the same time, to avoid constraining the filters to any particular class, thus enabling more expressive filters. The proposed Rotation Equivariant Vector Field Network (RotEqNet, \citet{marcos2016rotation}) achieves model-based invariance while reducing the number of required parameters by around one order of magnitude. This is done by sharing the same convolutional filter wights across all angles, thus providing regularization to irrelevant modes of variations \citep{leen1995neucomp}. 

In addition, the decrease in sample complexity allows models to be trained more efficiently. In this paper, we also analyze the effect that the amount of available ground truth has on the performance of CNNs learning for semantic labeling of overhead imagery based on two public datasets at submetric resolution: the Vaihingen and the Zeebruges benchmarks (see Section~\ref{sec:datasetup}).

In Section~\ref{sec:method} we briefly present the intuition behind RotEqNet, as well as its main components. In Section~\ref{sec:datasetup} we present the data and the setup of the experiments presented and discussed in Section~\ref{sec:res}.



\section{Rotation Equivariant Vector Field Networks (RotEqNet)}\label{sec:method}

In this paper, we propose to make a CNN equivariant to rotations by rotating filters and considering only the maximal activation across rotations. This section first recalls basics about CNNs and then presents the RotEqNet modules as a way of extending any CNN architecture into a rotation equivariant one. For more details about RotEqNet, the reader can refer to~\citet{marcos2016rotation}, where the theory was originally presented by the authors. 

\subsection{Convolutional neural networks for semantic labeling}
In this section we briefly present the building blocks of CNNs, as well as an example of a fully convolutional architecture to perform semantic labeling.

\subsubsection{CNN building blocks}
\begin{figure}
\includegraphics[width=\linewidth]{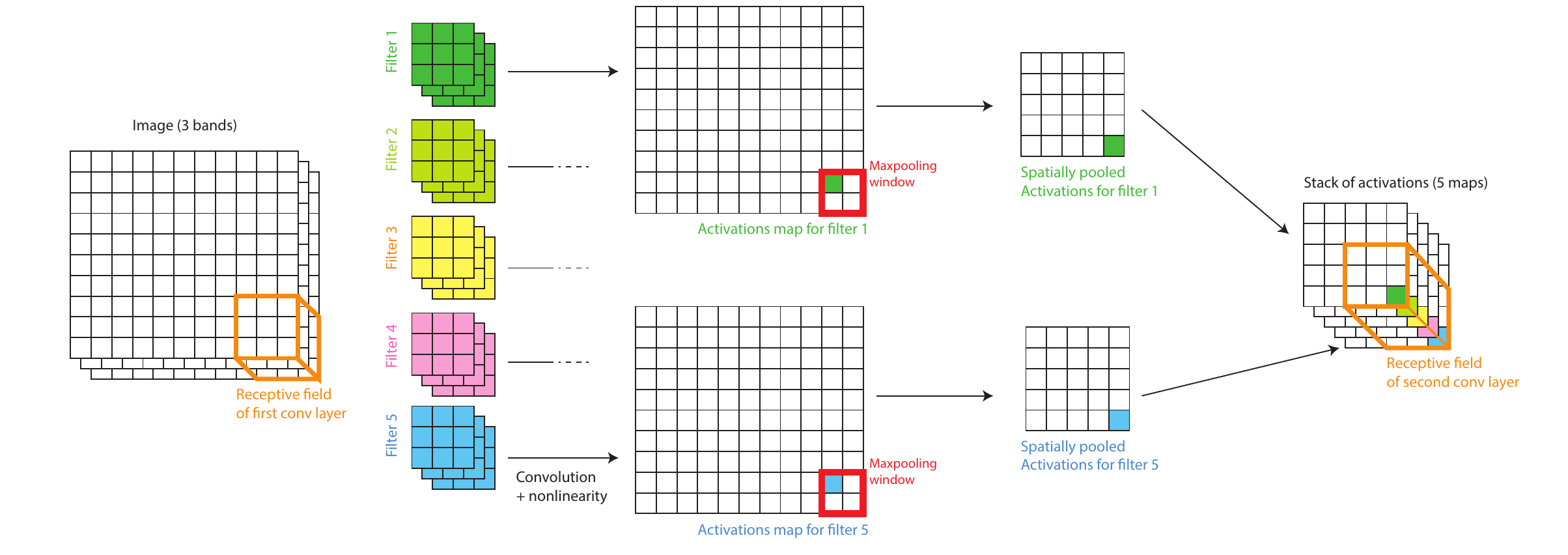}
\caption{Schematic of the first convolutional layer of a CNN. This layer learns Nf = 5 filters of size $3 \times 3 \times 3$ and applies a spatial pooling halving the size of the activation map (only two out of the five activation maps are shown for clarity). In the activation maps, the colored pixels (in green or blue) correspond to those receiving information from the receptive field marked in orange in the input image (left).}\label{fig:conv}
\end{figure}

CNNs consist of a cascade of operations applied to an input image $\textbf{x}\in \mathbb{R}^{M\times N\times d}$ such that it can be nonlinearly transformed in the desired output. The number of such operations defines the \emph{depth} of the network. CNNs are often organized in convolutional blocks as the one depicted in Fig.~\ref{fig:conv}.

The convolution operator 
\begin{equation}
\textbf{y}=\mathbf{x}\ast\mathbf{w}+b, \label{eq:conv}
\end{equation}
between $\mathbf{x}$ and a filter $\mathbf{w}\in \mathbb{R}^{m\times m\times d}$, where $b\in\mathbb{R}$ is the bias, produces a feature map $\mathbf{y}\in \mathbb{R}^{M-m+1\times N-m+1}$ by applying locally a scalar product operation between $\mathbf{w}$ and every patch in $\mathbf{x}$ of size $m \times m$ in a sliding window manner. 
A convolution block in a CNN corresponds to the convolution of the image with a series of filters, which are represented in different colors in Fig.~\ref{fig:conv}. 


The dimensionality of the activations equals the number of filters in the layer. To control the spatial extent of the activations after convolutions, it is common to apply zero-padding, which does not influence the value of the activations, but does compensate for the amount of pixels lost at the borders of the image. In order to obtain an advantage from the depth of the model, in terms of expressive power, it is necessary to apply some non-linear operation to the output of each convolution. The most common is the rectified linear unit (ReLU), which clips the negative values to zero, as $y = \max(0,x)$.


Once the activations are obtained, they are often pooled in the spatial domain, for example by taking the maximum value occurring in a very small (usually 2$\times$ 2) local window. This operation, called \emph{max-pooling} is represented in Fig.~\ref{fig:conv} by the red squares and, besides the obvious effect of reducing the amount of data, also allows the filters in the next layer to `see' more context of the original image: looking again at the schematic in Fig.~\ref{fig:conv}, if the first filters see a $3\times 3$ region, those of the second layer (orange cube on the right) will see a $3\times 3$ region in the reduced activations map, which coresponds to a $7 \times 7$ region in the original image. By cascading several convolutional and max-pooling blocks, the network actually becomes aware of a wide context around the pixel being considered, while reducing the number of required learnable parameters, and provides invariance to local (at each layer) and global (at the network level) translations. The latter is evident in image classification problems: an image contains a cat independently of where it is located. For semantic labeling tasks, max-poolings have the effect of learning locally consistent and multi-scale filters.

Two other operators are often used to improve the learning process: batch normalization and dropout. The former normalizes each feature map within a batch to have zero mean and unit standard deviation. The latter sets  a certain proportion of randomly selected feature maps to zero during training, thus preventing the filters from depending too much on one another.

\subsubsection{From patch classification to (dense) semantic labeling}

\begin{figure}
\includegraphics[width=\linewidth]{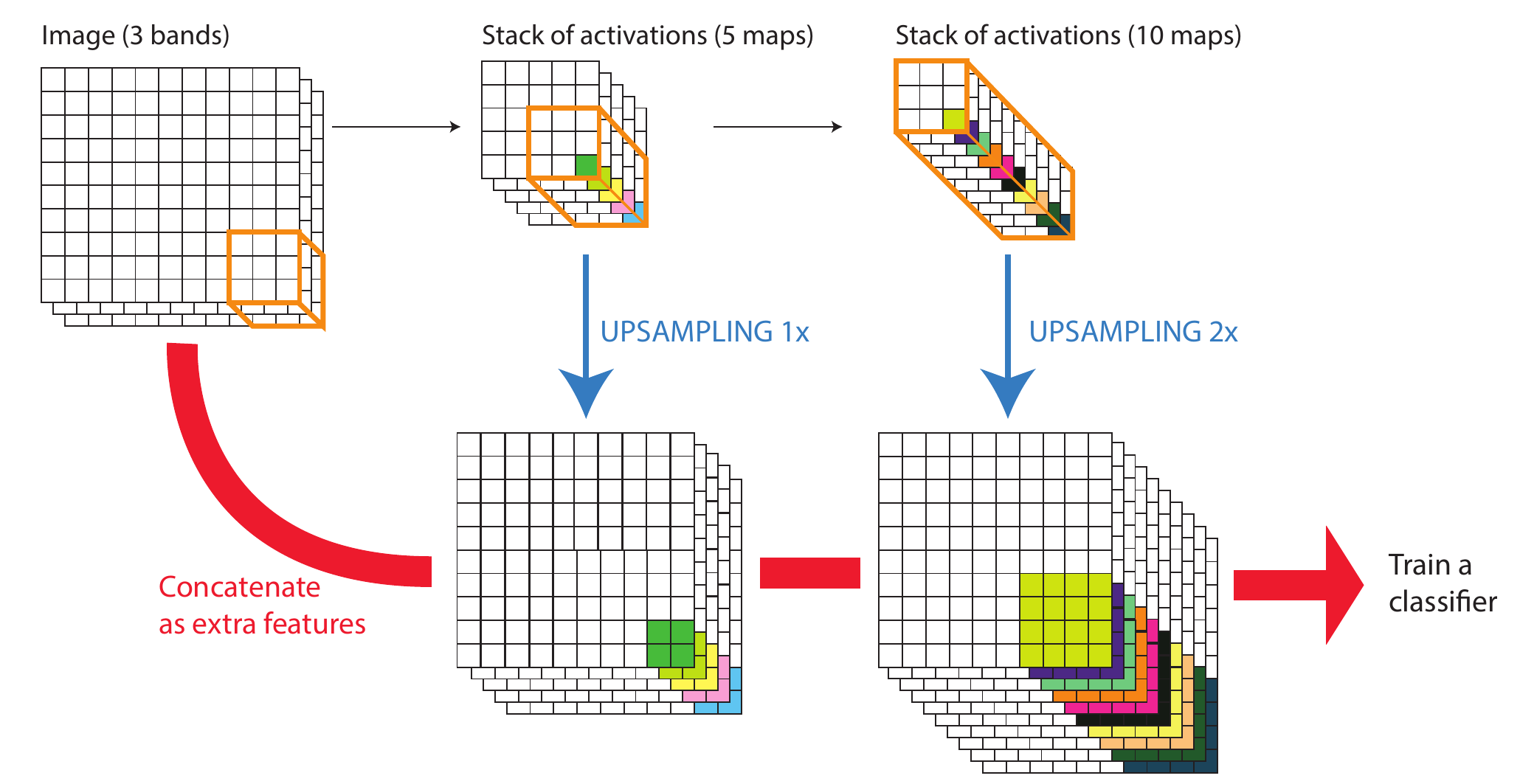}
\caption{Schematic of the model considered for dense semantic labeling. Each activation map in the CNN (top part) is upsampled at the original image resolution (blue arrows), concatenated to the original image (red arrow) and fed to a local fully connected layer ($1 \times 1$ convolutions), in this example using 18 features. 
}\label{fig:mlp}
\end{figure}

Early CNN models in vision were designed for tile (or image) classification, i.e. to provide a single label for an image. In semantic labeling, we are interested in obtaining a dense prediction, i.e. a map where each pixel is assigned to the  most likely label. As stated in the introduction, this can be achieved in a number of ways, including fully convolutional models~\citep{long2015cvpr}. A very simple way to perform dense predictions is to use the activation maps themselves as features to train a classifier predicting the label of every pixel. If max-pooling operations have been performed, spatial upsampling, e.g. by interpolation, is required to bring all activations to the same spatial resolution. One of these approaches, known as ``hypercolumns''~\citep{Hariharan_2015_CVPR}, using fixed upsamplings, is represented in Fig.~\ref{fig:mlp}. It follows the intuition that the different activation maps contain information about specific features extracted at different scales, from low-level ones (first layers react to corners, gradients, etc.) to more semantic and contextual ones (last layers activate to class-specific features). Therefore, a stack of such features can be used to learn an effective classifier for dense semantic labeling tasks, where spatial information is crucial. In remote sensing, the idea of hypercolumns was used in the architecture proposed by~\citet{Mag16}. In the experiments, we will use this architecture for dense semantic labeling, using two fully connected layers as classifier (see Section~\ref{sec:archi} for details), and train the model end-to-end.

\subsection{From translation to rotation invariance}
As mentioned above, CNNs are translation equivariant by design. To understand why it is more complex to achieve natural equivariance to rotations by means of convolutions, we will briefly summarize how translation equivariance is obtained by standard CNNs before moving to rotation equivariance and our proposed solution, RotEqNet.

\subsubsection{Translation equivariance in CNNs}
The convolution of an image $\mathbf{x}$ with a filter $\mathbf{w}$ (Eq.~(\ref{eq:conv})) is computed by applying the same dot product operation over all overlapping $m\times m$ windows on $\mathbf{x}$. If $\mathbf{x}$ is shifted by an integer translation in the horizontal and vertical directions, given a reference location, the same neighborhoods in $\mathbf{x}$ will exist in the translated $\mathbf{x}$. 
The corresponding convolution output, except for some possible border effects, are exactly the same up to some global translation constant. For this reason, neighborhood-based operations are translation equivariant when applied to images. The fact that the operation is local and produces a single scalar per neighborhood has another advantageous effect: the output can be effortlessly re-arranged in useful ways. Typically, the spatial structure of the activations is set to match the one of the input (as in CNNs, see Fig.~\ref{fig:conv}).


\subsubsection{Rotation equivariance in RotEqNet}

If we want the operator to be equivariant to rotation, the structure of the layer activations becomes more complex. One possibility would be to return a series of values corresponding to the convolution of $\mathbf{x}$ with rotated versions of the canonical filter $\mathbf{w}$. In this case, the activations $\mathbf{y}$ would be a 3D tensor where a translation in the $3^{rd}$ dimension corresponds to a rotation of $\mathbf{w}$. The covariance achieved in this way could easily be transformed into equivariance by means of pooling across orientations, since the value returned at each image location will remain constant when the image is rotated and thus a rotation of the input image will result in the same rotation of the output feature map.

\begin{figure}
\includegraphics[width=\linewidth]{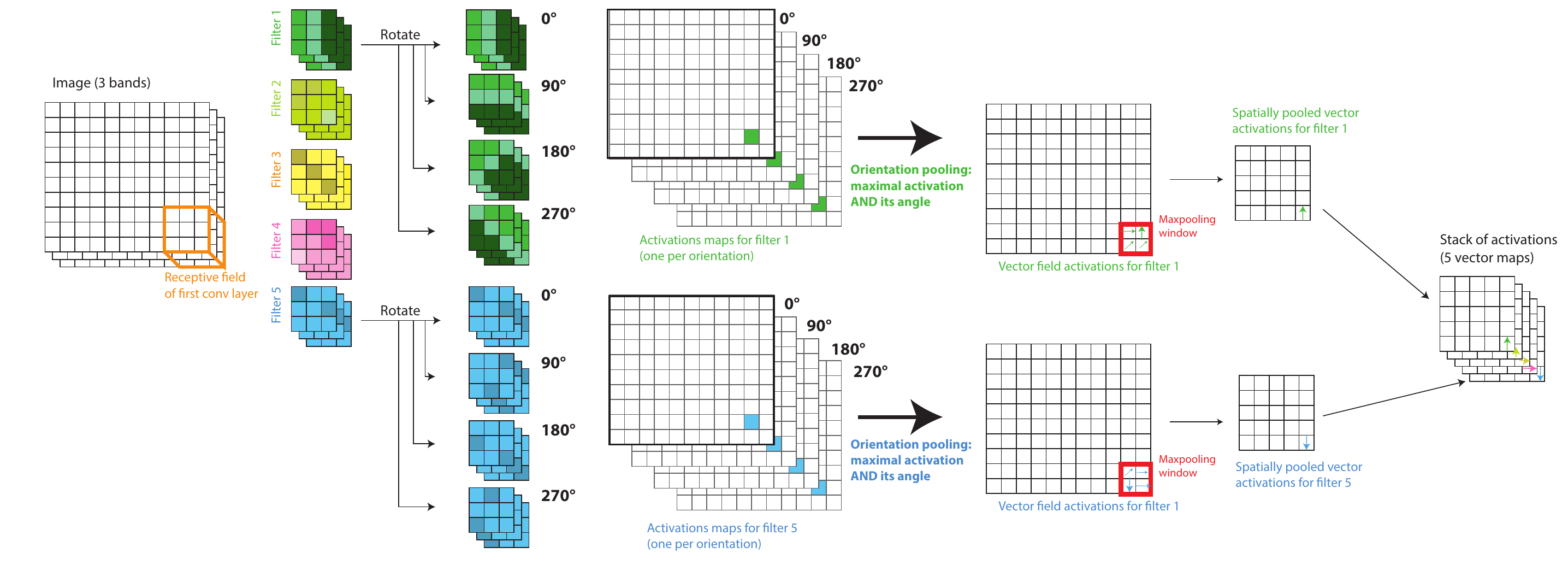}
\caption{Schematic of the first convolutional layer of RotEqNet. This layer learns Nf = 5 filters of size $3 \times 3 \times 3$. Each filter is rotated to a pre-defined range of angles and the activation at each orientation is computed. Then an orientation pooling retains only the maximal activation and the angle that generated it, thus providing a vector activation per pixel (represented by colored arrows). This vector map is pooled spatially as for conventional CNNs. The output is a stack of vector activations. In the activation maps, the colored pixels (in green or blue) correspond to those receiving information form the receptive field marked in orange in the input image (left). For clarity, only two out of the five activation maps are shown.}\label{fig:rotconv}
\end{figure}

In particular, we propose to perform a single-binned max-pooling operation across the newly added orientation dimension. At each location, it fires on the largest activation across orientations, returning its value (magnitude) \emph{and} the angle at which it occurred. This way, we are able to keep the 2D arrangement of the image and activations throughout the CNN layers, while achieving rotation equivariance as provided by this pooling strategy. Furthermore, this strategy allows the network to make use of the information about the orientation of feature activations observed in previous layers. Similar to spatial max-pooling, this \emph{orientation pooling} propagates only information about the maximal activation, discarding all information about non-maximal activations. This has the drawback of potentially loosing useful information (e.g. when two orientations are equally discriminant), but offers the advantage of reducing the memory requirements of both the model and the feature maps along the network, making them independent of the number of rotations used. Since the result of such pooling is no longer a scalar as in conventional CNNs, but a 2D vector (magnitude and angle), each activation map can now be treated as a vector field. Figure~\ref{fig:rotconv} schematizes this intuition and shows a RotEqNet convolutional block in comparison to the standard CNN convolutional block of Fig.~\ref{fig:conv}. 

\subsection{RotEqNet modules}
RotEqNet essentially involves rotating CNN filters and pooling across the orientation space to retrieve the maximal activations and their angle observed at each location and per filter. To achieve such behavior, several building blocks of CNNs must be re-designed in order to accommodate vector field inputs / outputs. In this section, we briefly summarize how the convolution and pooling operators have been modified. Modifications of spatial pooling and batch normalization are straightforward and we invite the interested reader to consult~\citet{marcos2016rotation} for more details.

\subsubsection{Rotating convolution}
As introduced above, rotation equivariance can be achieved by computing the activations on a series of rotated versions of the filters being learned. This boils down to calculate rotated versions of each main (or \emph{canonical}) filter at $R$ orientations $\boldsymbol{\alpha} = [\alpha_1, \ldots, \alpha_{R}]$. In case of remote sensing images, for which the orientation might be completely arbitrary, $\boldsymbol{\alpha}$ can span the entire $360^\circ$ rotation space, while in other applications with a clear top-down relations, one could limit the angles to a smaller range (e.g. it is unlikely that a tree depicted in a ground level image is oriented in the left-right direction, but some tilts due to camera shake could be present).

The rotation of the filter is obtained by resampling $\mathbf{w}$ with bilinear interpolation, after rotation of $\alpha_r$ degrees around the filter center. The position $[i',j']$ after rotation of a specific filter weight, originally located at $[i,j]$ in the canonical form, is
\begin{equation}
[i',j'] = [i,j]
\bigg[
\begin{array}{c c}
\cos(\alpha_r) & \sin(\alpha_r)\\
-\sin(\alpha_r) & \cos(\alpha_r)
\end{array}
\bigg].
\end{equation}
Coordinates are relative to the center of the filter. Since the rotation can force weights near the corners of the filter to be relocated outside of its spatial support, only the weights within a circle of diameter $m$ pixels, the filter size, are used to compute the convolutions. The output tensor for filter $\mathbf{w}$, of size $\mathbf{y}\in \mathbb{R}^{M\times N\times R}$, consists of $R$ feature maps (see the center part of Fig.~\ref{fig:rotconv}), each one computed as 
\begin{equation}
\mathbf{y}^{(r)} = \mathbf{x}\ast\mathbf{w}_{r} + b \quad \forall r = 1,2\dots R,
\end{equation}
where $(\ast)$ is a standard convolution operator, and $b$ is a shared bias  across all rotations. 
%
%
The tensor $\mathbf{y}$ encodes the roto-translation output space such that rotation of the input corresponds to a translation across the feature maps. Only the canonical, non rotated, version of $\mathbf{w}$ is actually stored in the model. During backpropagation, gradients flow through the filter with maximal activation, very similarly to the max-pooling case. Consequently, the gradients have to be aligned to the rotation of the canonical filter, which is recovered thanks to the angle as given by the orientation pooling. Thus, filters are updated as:
\begin{equation}
\nabla \mathbf{w} = \sum_r \mathtt{rotate}(\nabla\mathbf{w}_{r},-\alpha_r),
\end{equation}
%


\subsubsection{Orientation pooling}
The rotating convolution above outputs a set of $R$ activations per canonical filter, each one corresponding to one rotated version of $\mathbf{w}$. To avoid the explosion of dimensionality related to the propagation of all these activations to the next layer, we perform a pooling across the space of orientation aiming as pushing forward only the information relative to the direction of maximal activation. In order to preserve as much information as possible, we keep two kinds of information: the \emph{magnitude of activation} and the \emph{orientation that generated it}.

To do so, we extract a 2D map of the largest activations $\boldsymbol{\rho} \in \mathbb{R}^{M\times N}$ and their corresponding orientations $\boldsymbol{\theta} \in \mathbb{R}^{M\times N}$. Specifically, for activations located at $[i,j]$ we get:
\begin{equation}
{\rho}[i,j] =  \max_{r} {y}[i,j,r], \\
\end{equation}

\begin{equation}
{\theta}[i,j] = \frac{360}{R}\, \arg\max_{r} {y}[i,j,r].
\end{equation}

This can be treated as a polar representation of a 2D vector field as long as ${\rho}[i,j] \ge 0 \quad \forall i,j$. This condition is met when using a function on $\mathbf{y}$ that returns non-negative values: we therefore employ the Rectified Linear Unit (ReLu) operation, defined as $\texttt{ReLu}({\rho})=\max({\rho},0)$. In the backward pass the magnitude of the incoming 2D vector gradient is passed to the corresponding maximally activated position, ${y}[i,j,r_{max}]$, as is done with standard max-pooling.


\subsubsection{Dealing with vector inputs}

Note that the orientation pooling block outputs vector fields, where each location in the activation carries both the maximal magnitude and its orientation observed in polar representation (see the rightmost matrix in Fig.~\ref{fig:rotconv}). This means that the output of such pooling is vectorial and cannot be used anymore in a traditional convolutional layer. However, if we convert this polar representation into Cartesian coordinates, each filter $\mathbf{w}$ produces a vector field feature map $\mathbf{z}\in \mathbb{R}^{M\times N \times 2}$, where the output of each location consists of two values $[u,v]\in \mathbb{R}^2$ encoding the same information. 

\begin{equation}
\mathbf{u} = \texttt{ReLu}(\boldsymbol{\rho}) \cos(\boldsymbol{\theta})
\end{equation}

\begin{equation}
\mathbf{v} = \texttt{ReLu}(\boldsymbol{\rho}) \sin(\boldsymbol{\theta})
\end{equation}

Since the horizontal and vertical components $[u,v]$ are orthogonal, the convolution of two vector fields can be computed summing standard convolutions calculated separately in each component:
\begin{equation}
(\mathbf{z}\ast\mathbf{w}) = (\mathbf{z}_u\ast\mathbf{w}_u) + (\mathbf{z}_v\ast\mathbf{w}_v),\label{eq:orth}
\end{equation}

By using this trick, we can now calculate convolutions between vector fields and design deep architectures which are rotation equivariant.

\section{Data and setup}\label{sec:datasetup}

\subsection{Datasets}
We test the proposed system on two recent benchmarks that raised significant interest thanks to the dense ground truth provided over a set of sub-decimeter resolution image tiles: the Vaihingen and Zeebruges data, which are briefly described below. Both datasets consist of three optical bands and a Digital Surface Model (DSM). Since since using the DSM has been shown to improve the segmentation results~(\cite{audebert2017fusion,marmanis2016semantic,volpi2017dense}) we use it in all of our experiments. We do this by stacking the DSM with the optical data and treating it as an additional band, as in~\cite{volpi2017dense}, since this has almost no impact in the total number of model parameters.

\subsubsection{Vaihingen benchmark}
The Vaihingen benchmark dataset has been provided to the community as a challenge organized by the International Society for Photogrammetry and Remote Sensing (ISPRS) Commission III.4, the ``2D semantic labeling contest''\footnote{\url{http://www2.isprs.org/commissions/comm3/wg4/semantic-labeling.html}}. The dataset is composed of 33 orthorectified image tiles acquired over the town of Vaihingen (Germany), with an average size of $2494 \times 2064$ and a spatial resolution of 9 cm. Among the 33 frames, 16 are fully annotated and distributed to participants, while the remaining ones compose the test set and their ground truth is not distributed. Images are composed by 3 channels: near infrared (NIR), red (R) and green (G). The challenge also provides a DSM coregistered to the image tiles. We use a normalized version of the DSM (nDSM), where the heights are relative to the nearest ground pixel, redistributed by~\citet{gerke2015techrepo}. One of the training tiles is illustrated in Fig.~\ref{fig:V}.

The task involves 6 land-cover / land-use classification classes: ``impervious surfaces'' (roads, concrete flat surfaces), ``buildings'', ``low vegetation'', ``trees'', ``cars'' and a class of ``clutter'' to group uncategorized surfaces and noisy structures.  Classes are highly imbalanced, with the classes ``buildings'' and ``impervious surfaces'' accounting for roughly 50\% of the data, while classes such as ``car'' and ``clutter'' account only for 2\% of the total labels. 

In our setup, 11 of the 16 fully annotated image tiles are used for training, and the remaining ones (tile ID 11, 15, 28, 30, 34) for testing, as in~\citet{sherrah2016arxiv,volpi2017dense,maggiori2017convolutional}.


\begin{figure}
\begin{tabular}{ccc}
\includegraphics[width=.3\linewidth]{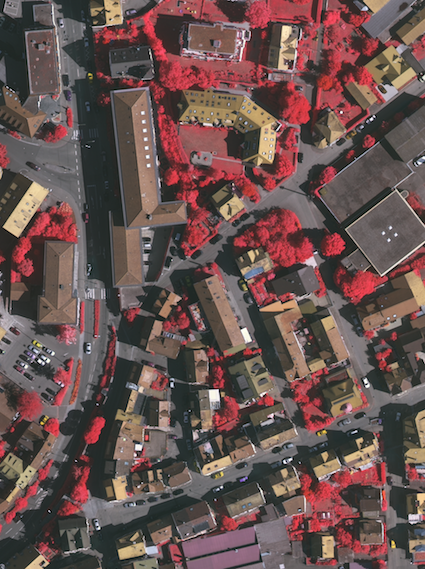}&
\includegraphics[width=.3\linewidth]{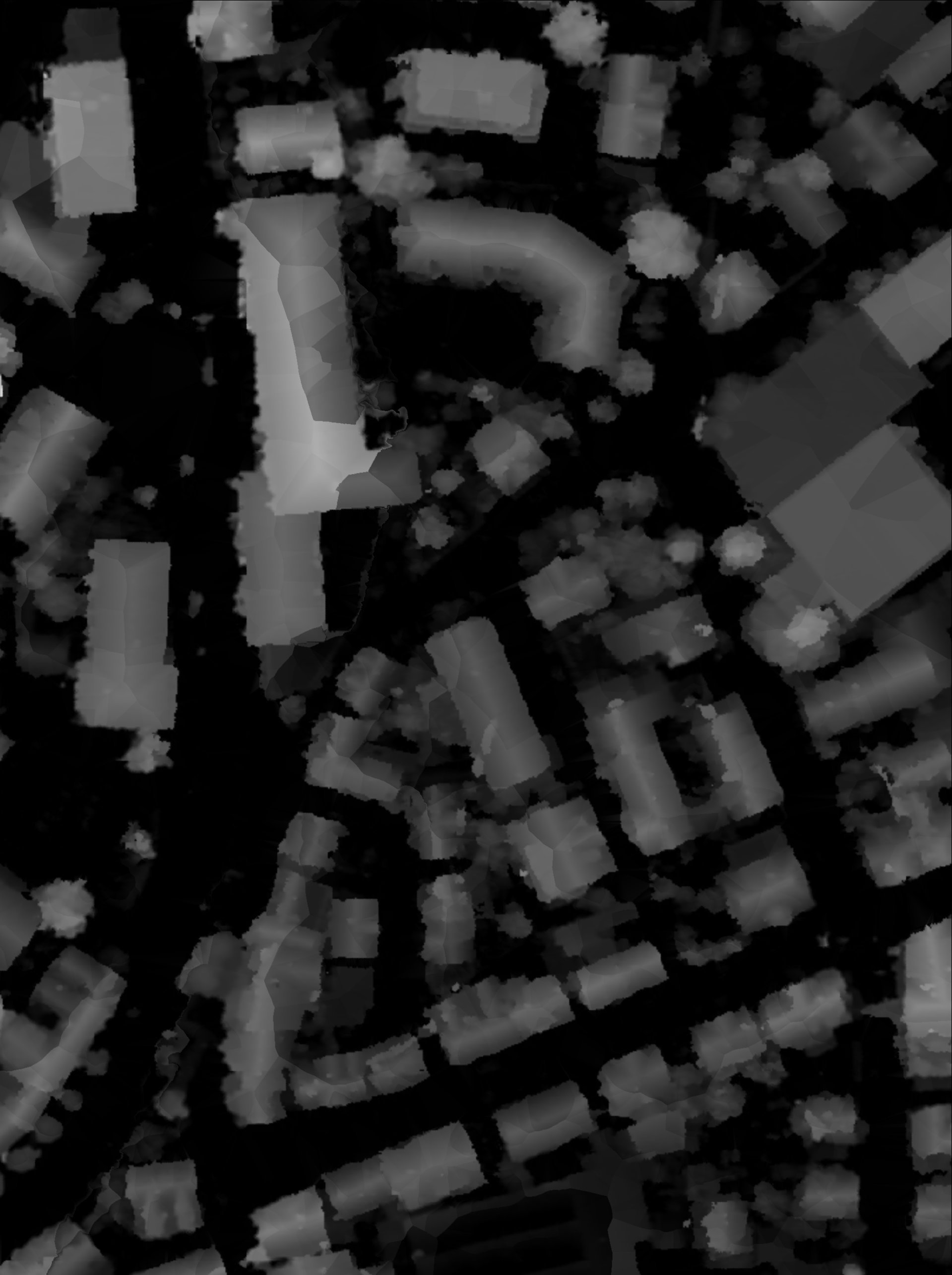}&
\includegraphics[width=.3\linewidth]{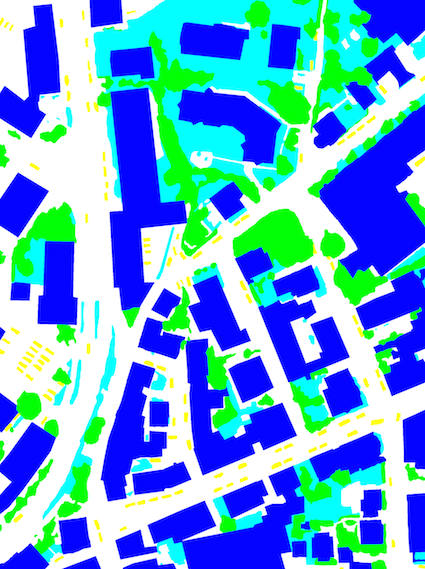}\\
Image tile \#1 & nDSM & ground truth \\
\end{tabular}
\caption{one of the training tiles of the Vaihingen dataset: left: image; center: nDSM; right: ground truth.}\label{fig:V}
\end{figure}

\subsubsection{Zeebruges benchmark}\label{sec:Zeebruges}

This benchmark has been acquired in 2011 over the city of Zeebruges (Belgium) and it is has been provided as part of the IEEE GRSS Data Fusion Contest in 2015 \citep{DFCA}\footnote{\url{http://www.grss-ieee.org/community/technical-committees/data-fusion/2015-ieee-grss-data-fusion-contest/}}. It is composed by seven  tiles of $10000 \times 10000$ pixels. The tiles have a spatial-resolution of 5 cm and represent RGB channels only. Five of the seven images are released with labels~\citep{Lag15} and used for training, while the remaining two are kept for testing the generalization accuracy, accordingly to the challenge guidelines. This dataset also comes with a Lidar point cloud, that we processed into a DSM by averaging point clouds locally and interpolating where necessary.

The semantic labeling problem involves 8 classes, as proposed in~\citet{Lag15}: the same six as in the Vaihingen benchmark, plus a  ``water'' and ``boats'' class. It is  worth mentioning that, since a large portion of the are is covered by a harbour, most of the structures and cargo containers are labeled as ``clutter''. Another major difference to the Vaihingen dataset is that the ``Water'' class is predominant, as it represents 30\% of the training data, while cars and boats together count just a mere 1\%. Also, 1\% of the data belongs to an ``uncategorized'' class, which is not accounted for in the labeling problem. The lack of a NIR channel and a higher sample diversity make this benchmark more challenging than the previous one, as we will see in Section~\ref{sec:res}.


\begin{figure}
\begin{tabular}{ccc}
\includegraphics[width=.3\linewidth]{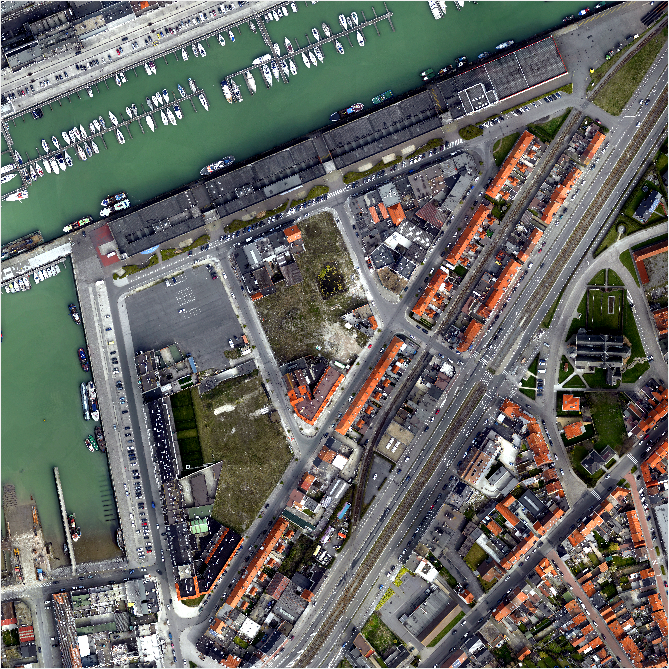}&
\includegraphics[width=.3\linewidth]{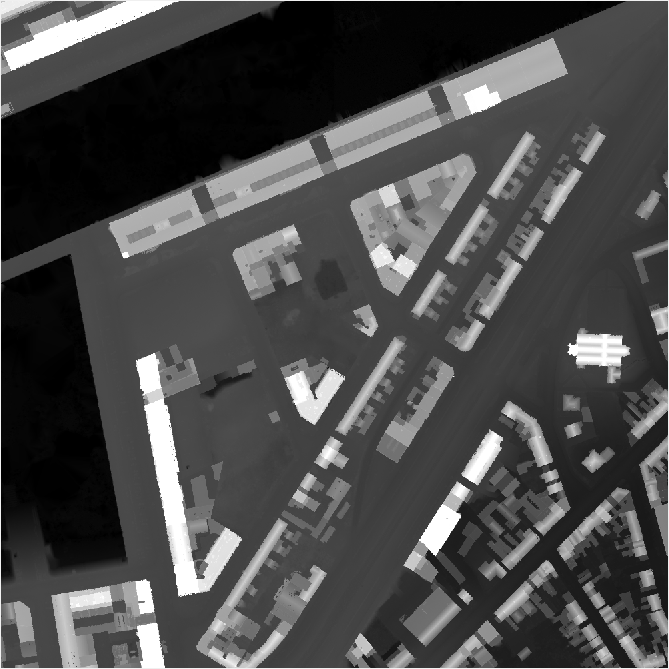}&
\includegraphics[width=.3\linewidth]{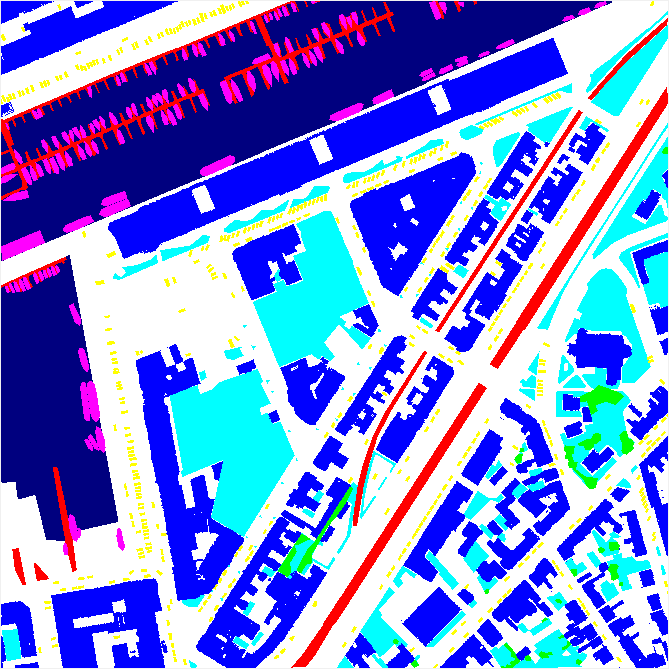}\\
Image tile RGB & nDSM & ground truth \\
\end{tabular}
\caption{one of the training tiles of the Zeebruges dataset: left: image; center: nDSM; right: ground truth.}\label{fig:V}
\end{figure}

\subsection{Experimental setup}\label{sec:exp}

\subsubsection{CNN architecture}\label{sec:archi}

\begin{figure}[!t]
\includegraphics[width=\linewidth]{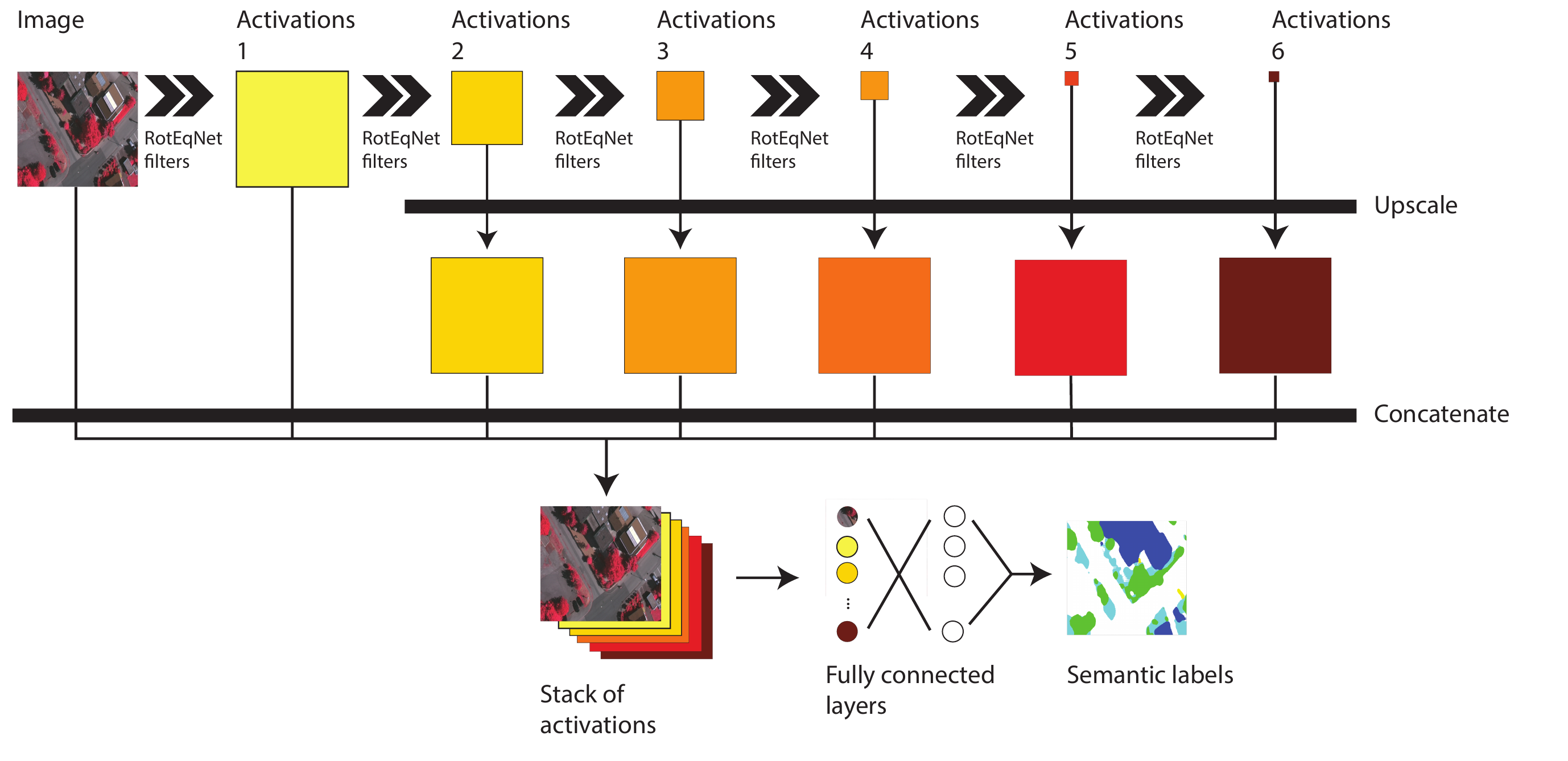}
\caption{Hypercolumn based architecture used in all our experiments. Note that all the layers are rotation equivariant, since all the convolutional layers are either RotEqNet convolutions or fully connected ($1\times 1$) standard convolutions, which are also rotation equivariant by construction.}
\label{fig:arch}
\end{figure}

We use a RotEqNet architecture based on hypercolumns~\citep{Hariharan_2015_CVPR}, in which every convolutional layer before the concatenation is a rotating convolution. After the concatenation, only standard fully connected layers are used, since $1\times 1$ convolutions are inherently rotation equivariant. See Fig.~\ref{fig:arch} for a schematic of the full architecture used.
In both experiments we build the baseline CNN with the exact same architecture as its RotEqNet counterpart, but with four fold more filters in each layer, resulting in approximately 10 fold more parameters. At this model size the performance started to saturate.

We use an architecture with six convolutional layers, each with downsampling by a factor of 2 using max-pooling. The number of filters per layer in each of the convolutional layers is set as $[2, 2, 3, 4, 4, 4]\cdot\textrm{Nf}$, where the $i^{th}$ element of the vector represents the number of filters in the $i^{th}$ layer, such that $\textrm{Nf}$ is the only parameter used to change the size of the models. All the convolutional layers use $7\times 7$ filters. This size allows to capture oriented patterns in the corresponding image or feature map, as seen in Fig.~\ref{fig:filts}. After applying a linear rectification (ReLU) and batch normalization, each activation map is then upsampled to the size of the original image using bilinear interpolation and concatenated, together with the raw image (see bottom part of Fig.~\ref{fig:mlp}), and processed with three layers of $1\times 1$ convolutions with $[50\cdot\textrm{Nf},50\cdot\textrm{Nf},C]$ filters, where $C$ is the number of classes. This is followed by a softmax normalization. The $1\times 1$ convolutions implement a local fully connected layer, or, in other terms, performs local classification by a multi-layer perceptron (MLP). These $1\times1$ convolutions are inherently rotation equivariant, so we use standard convolutions as in~\citet{long2015cvpr}. The whole pipeline is learned jointly end-to-end. 


Our model performs arbitrarily dense prediction, i.e. given an arbitrarily sized input patch the output will always be a prediction map with the same size. As a result, the CNN architecture is fixed and the dataset reshaped to a series of fixed-sized patches. For the Vaihingen dataset, the spatial extent of the inputs is $512 \times 512$, while for Zeebruges is $500\times 500$. Note that the input size does not influence the results.   

All models are trained with stochastic gradient descent with momentum (fixed to 0.9), while other hyperparameters are tuned by minimizing validation errors on $30$ samples randomly selected from the training set. The batch size is $4$ in RotEqNet and $2$ in the standard CNN, because of the latter's higher memory requirements. In both benchmarks, we perform data augmentation consisting of random rotations, uniformly sampled between $0^o$ and $360^o$, and randomly flipping the tiles in the vertical or horizontal dimension. Note that performing full $360^o$ rotations for data augmentation is not strictly necessary when using RotEqNet, but it has been shown to be additionally improve the performance (see results in~\citet{marcos2016rotation}) and doing so makes a comparison with standard CNNs easier, since they are trained under more similar conditions. 
As will be discussed below, data augmentation is required by standard CNNs in order to be able learn rotation invariance from examples, given that enough filters can be learned. Regarding RotEqNet, rotating inputs does not have a direct effect on learning diverse filters, but rather on data interpolation making a same input tile looking different numerically (an effect also improving training for standard CNNs). 

All models are trained from scratch and filters are initialized using the improved Xavier method.

The hardware used in all experiments consists of a single desktop with 32 GB of RAM and a 12 GB Nvidia Titan X GPU.

\subsection{Experimental Setup}

\subsection{Vaihingen}
In the case of Vaihingen, we report a comparison between RotEqNet and standard CNNs trained without rotating convolutions. In order to test the sensitivity to the amount of ground truth, we train three models per architecture, using respectively 4\%, 12\% and 100\% of the available training set. We compare architectures with the same structure and number of layers, only varying the number of filters. We compare a small RotEqNet model with a CNN of larger capacity (but no built-in rotation equivariance). The size of both models was chosen to be the smallest that would obtain over $87\%$ overall accuracy on the validation set, which is in line with the results published in~\cite{volpi2017dense}. The final number of filters defined in this way was found to be Nf$=3$ for RotEqNet ($\approx 10^5$ parameters) and Nf$=12$ for the standard CNN ($\approx10^6$ parameters). The models using the full dataset were trained for 22 epochs. In the RotEqNet models the learning rate was $2\cdot10^{-2}$ in the first 11 epochs, followed by six epochs at $4\cdot10^{-3}$ and five at $8\cdot10^{-4}$, while the weight decay was $4\cdot10^{-2}$, $4\cdot10^{-3}$ and $8\cdot10^{-4}$ respectively. In the standard models those values were halved. This difference is due to a larger number of gradient update iterations in the standard CNN caused by the need to use a smaller mini-batch due to the larger memory requirements. For the experiments with a reduced training set the number of epochs was increased such that all the models would see the same number of iterations (\emph{i.e.} mini-batches).
\subsection{Zeebruges}
In the case of Zeebruges, we compare RotEqNet with results from the literature~\citep{DFCA}, as they report results obtained with much larger architectures, both pre-trained or learned from scratch. Since this dataset is more complex than the previous one, we increased the model size and trained three RotEqNet models with Nf$=[4,5,7]$. The training schedule consisted of 34 epochs, with the first 12 epochs using a learning rate of $1\cdot10^{-2}$, 12 more with $2\cdot10^{-3}$ and 10 at $4\cdot10^{-4}$. The weight decay for the same segments was set to $6\cdot10^{-2}$, $1.2\cdot10^{-2}$ and $2.4\cdot10^{-3}$ respectively.



\section{Results and discussion}\label{sec:res}

\subsection{Vaihingen}
Table~\ref{tab:vai} shows the results in terms of the per class F1 scores, the overall accuracy (OA) and average accuracy (AA) for the experiments on the Vaihingen dataset. We observe that both models reach over 87\% OA when using the whole dataset, in line with recent publications and with the accuracy obtained by RotEqNet in the withhold test set, evaluated as 87.6\% by the benchmark server. This only drops to around 84.7\% when just 4\% of the training set is used, suggesting that this dataset is highly redundant.

\begin{table}[h]
	\centering
	\caption{Results on the Vaihingen validation set. F1 scores per class and global average (AA) and overall accuracies (OA). Best result per row is in dark gray, second in light gray.}
	\label{tab:vai}
	\begin{tabular}{c|ccc|ccc|c|c}
		\hline
		Model    	& \multicolumn{3}{c|}{RotEqNet} & \multicolumn{3}{c|}{CNN} & CNN-FPL$^*$ & ORN\\\hline\hline
		\# params. & \multicolumn{3}{c|}{$10^5$} & \multicolumn{3}{c|}{$10^6$} & $10^7$ & $10^5$\\
		\% train set &	4\%  & 12\% & 100\% & 4\%  & 12\% & 100\% & 100\% & 100\% \\ \hline\hline
		
		Impervious & 88.0 & 88.7 & 89.5\cellcolor[gray]{0.8} & 86.9 & 88.8 & \textbf{89.8}\cellcolor[gray]{0.6} & - & 88.1\\
		Buildings & 94.1 & 94.6 & 94.8\cellcolor[gray]{0.8} & 92.5 & 92.9 & 94.6 & - & \textbf{95.4}\cellcolor[gray]{0.6}\\
		Low veg. & 71.6 & 75.6 & \textbf{77.5}\cellcolor[gray]{0.6} & 74.5 & 74.5 & 76.8\cellcolor[gray]{0.8} & - & 70.9\\
		Trees & 82.3 & 85.6 & 86.5\cellcolor[gray]{0.8} & 83.3 & 84.4 & 86.0 & - & \textbf{92.1}\cellcolor[gray]{0.6}\\
		Cars & 62.7\cellcolor[gray]{0.8} & 62.5 & \textbf{72.6}\cellcolor[gray]{0.6} & 52.7 & 54.4 & 54.5 & - & 59.7\\
		\hline
		OA &  84.6 & 86.6 & 87.5\cellcolor[gray]{0.8} & 84.8 & 85.5 & 87.4 & \textbf{87.8}\cellcolor[gray]{0.6} & 87.0 \\ 
		AA & 78.4 & 80.5 & \textbf{83.9}\cellcolor[gray]{0.6} & 76.1 & 77.1 & 78.2 & 81.4\cellcolor[gray]{0.8} & 81.2 \\ 
		\hline
		\multicolumn{6}{l}{$^*$ = from \cite{volpi2017dense}}
	\end{tabular}
\end{table}

The advantage of using RotEqNet becomes more apparent when measuring the AA, mostly because of an improved accuracy detecting the car class. We hypothesize that RotEqNet might be better suited to detect objects with clear and consistent boundaries, such as cars, because it is being forced to learn oriented filters, better adapted to detect edge-like features. Surprisingly, RotEqNet improves its performance gap with respect to the standard CNN when the amount of available ground truth increases. This suggests that encoding rotation invariance allows the model to concentrate more on solving the semantic labeling task, rather than having to learn to be invariant to rotations.

As a comparison, we show the results recently published by \cite{volpi2017dense} using a much larger model and those obtained by applying the method by \cite{zhou2017oriented} (ORN) with a model of the same architecture and size as ours.

\begin{figure}
	\centering
	\begin{tabular}{cc}
		\includegraphics[width=.15\linewidth]{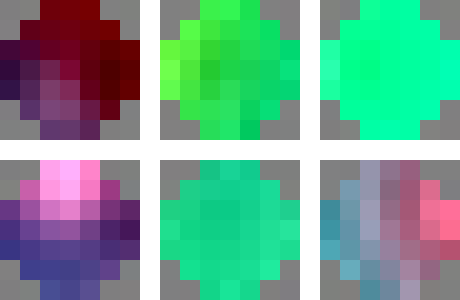}&
		\includegraphics[width=.15\linewidth]{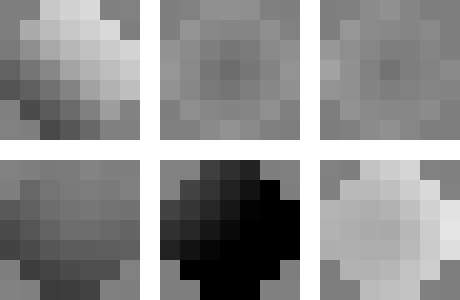}\\
		a & b \\
		\includegraphics[width=.3\linewidth]{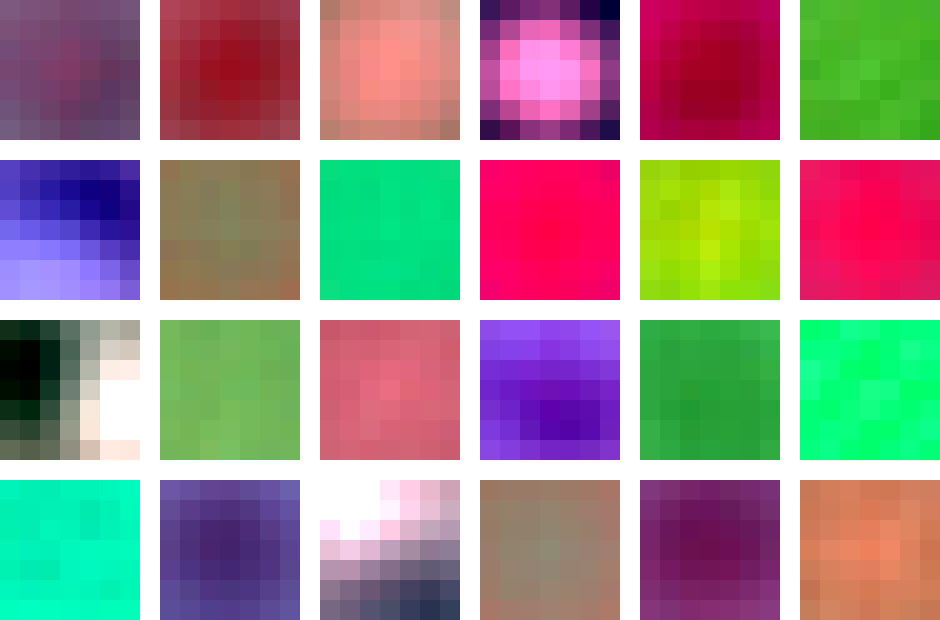}&
		\includegraphics[width=.3\linewidth]{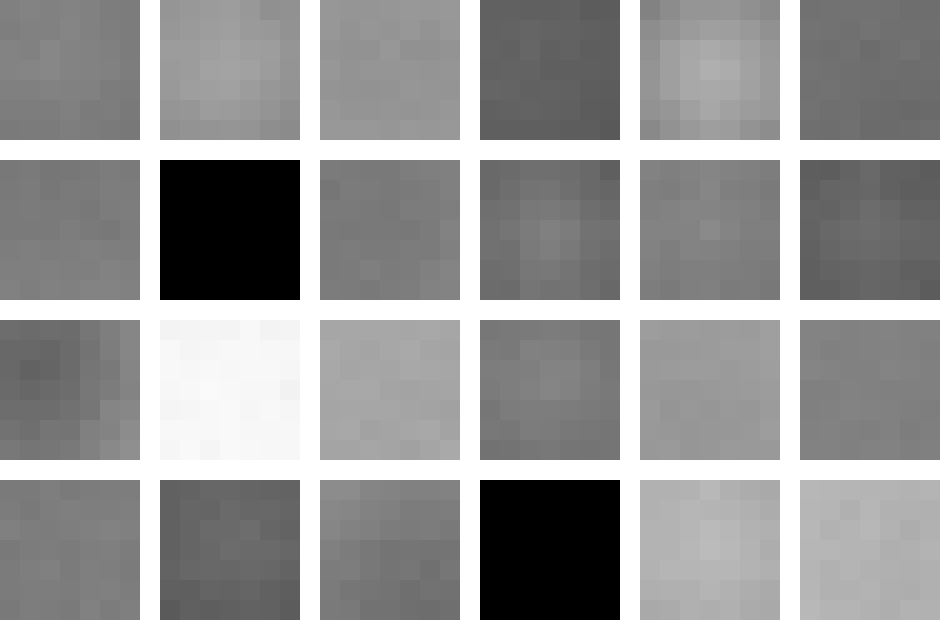}\\
		c & d \\
	\end{tabular}
	\caption{Visualization of all the filters learned on the first layer of the RotEqNet model on Vaihingen, a) on the optical channels, b) on the NDSM, and of the Standard CNN model, c) on the optical channels, d) on the NDSM. The filters are not normalized to appreciate the relative importance of each channel.}\label{fig:filts}
\end{figure}

In order to glimpse at what is being learned by both models, in Figure~\ref{fig:filts} we show all the $7\times 7$ filters learned in the first convolutional layer in each model. Note that the values near the corner of the filters in the RotEqNet model are zero because the support of the filter is a disk circumscribed in the $7\times 7$ grid. Out of the six filters learned by RotEqNet in the first layer, three seem to have specialized in learning corner features involving vegetation (the reddish tones means high response to the near infrared channel), one on low lying impervious surfaces and two in high impervious surfaces, which could be interpreted as rooftops. On the other hand, a majority of the standard filters seem to be relatively less structured and respond to some combination of color and height. We can also see a few instances of edge detectors that have been learned in different orientations. Note that the particular orientation of the RotEqNet filters is arbitrary and any other rotated version could have been learned as the canonical filter. 

RotEqNet does not need to learn filters that are rotated versions of each other because all of these versions are explored by applying each filter at many different orientations. This means that, while standard CNNs require data augmentation to perform well in a rotation equivariant setting, RotEqNet extracts features at different orientations and keeps the largest activations, effectively analyzing the input at different orientations without rotating it explicitly.

\begin{figure}
\begin{tabular}{ccccc}
Optical image & nDSM & GT & CNN & RotEqNet\\
\includegraphics[width=.17\linewidth]{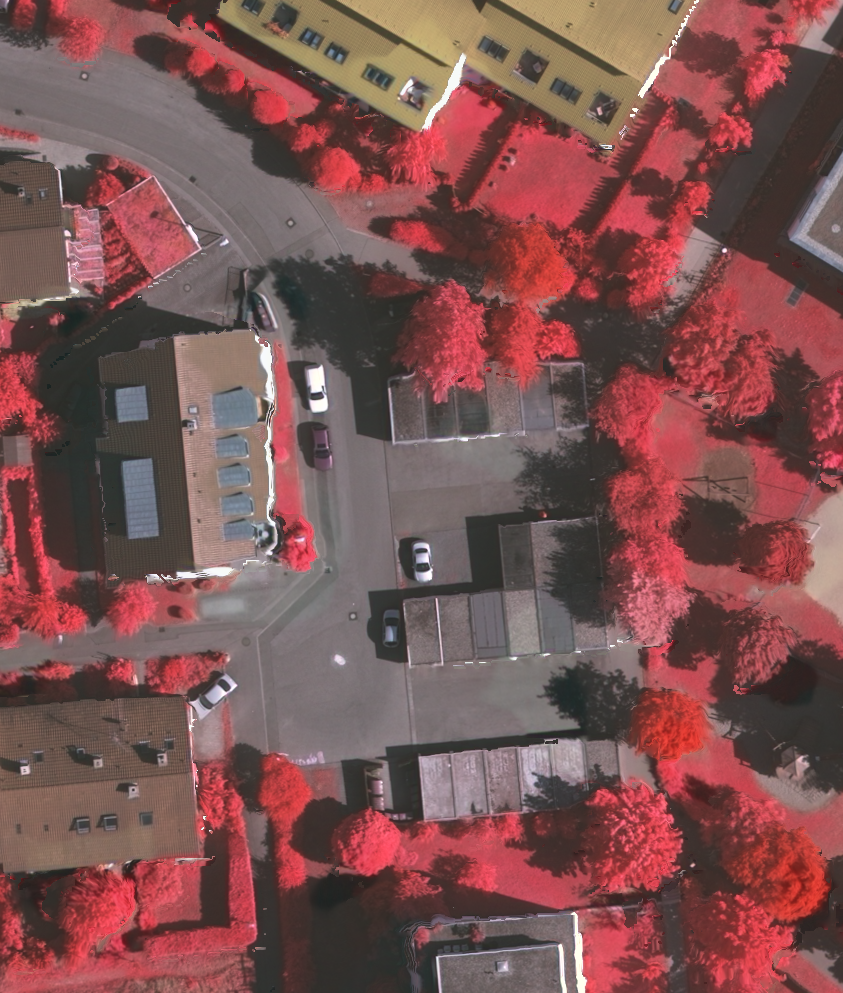} &
\includegraphics[width=.17\linewidth]{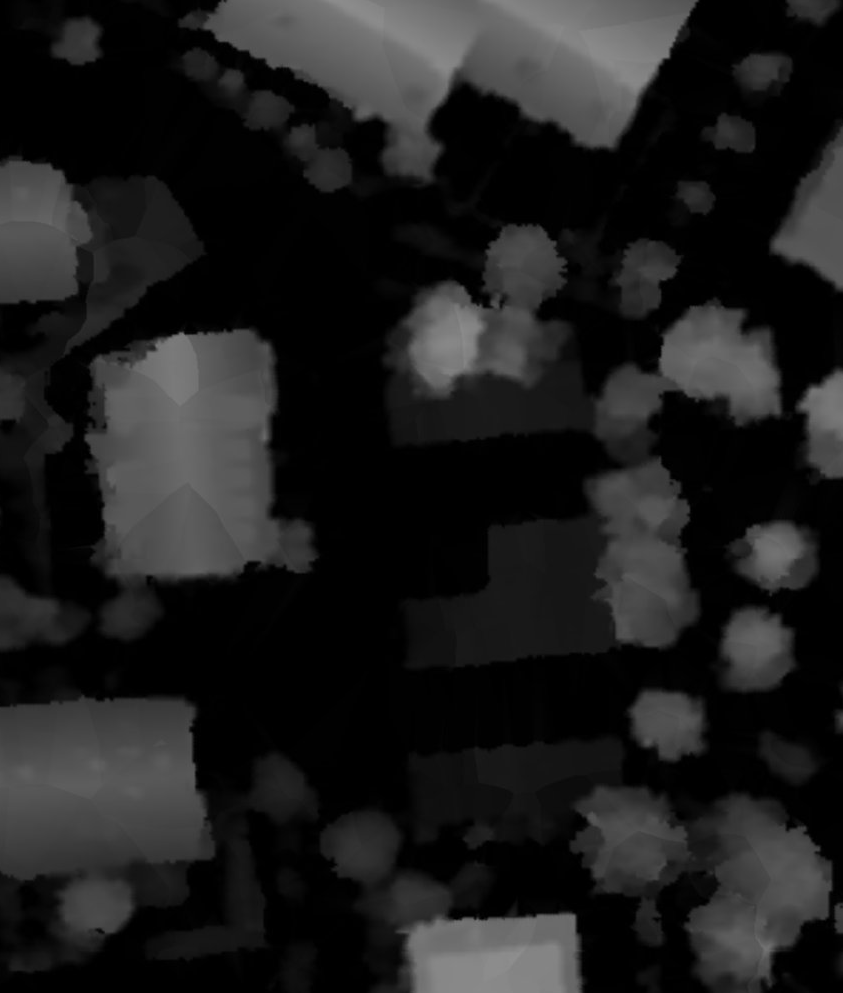} &
\includegraphics[width=.17\linewidth]{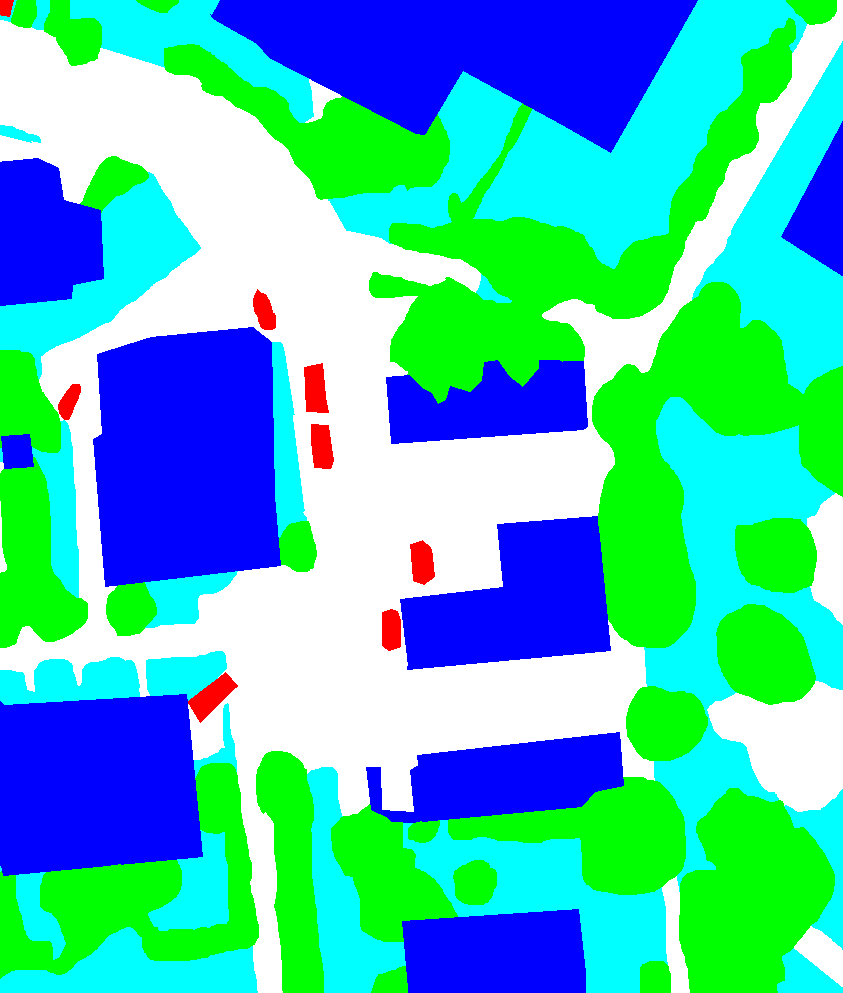} &
\includegraphics[width=.17\linewidth]{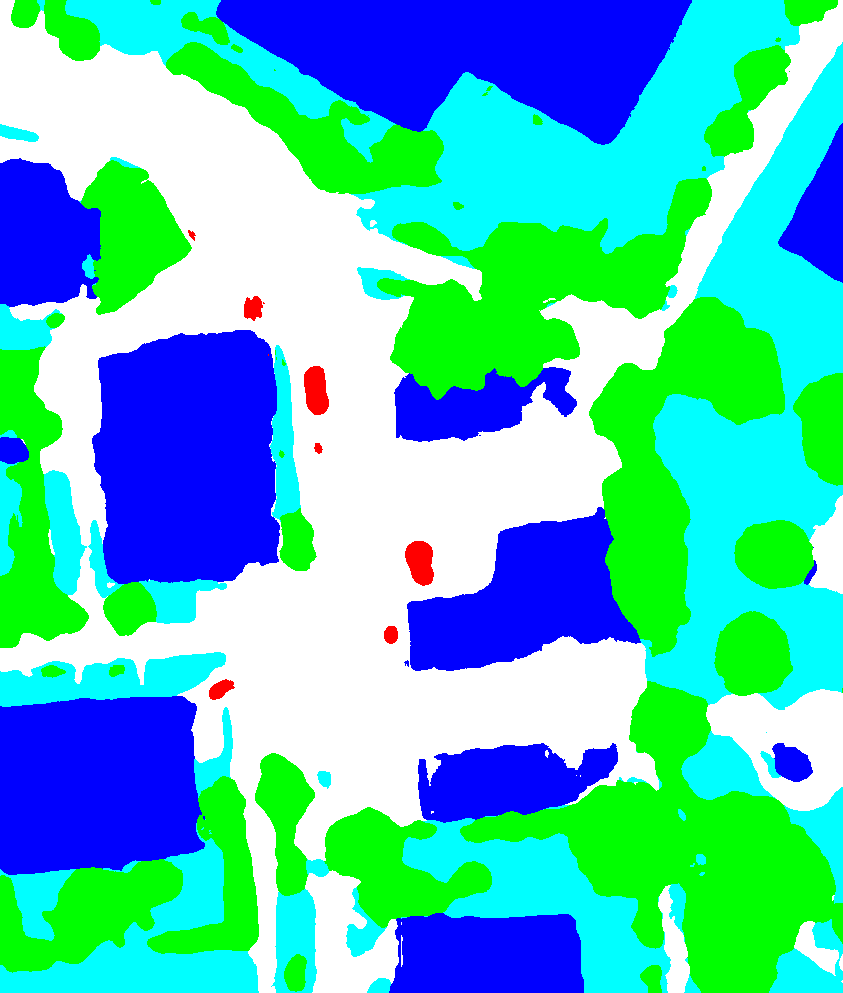} &
\includegraphics[width=.17\linewidth]{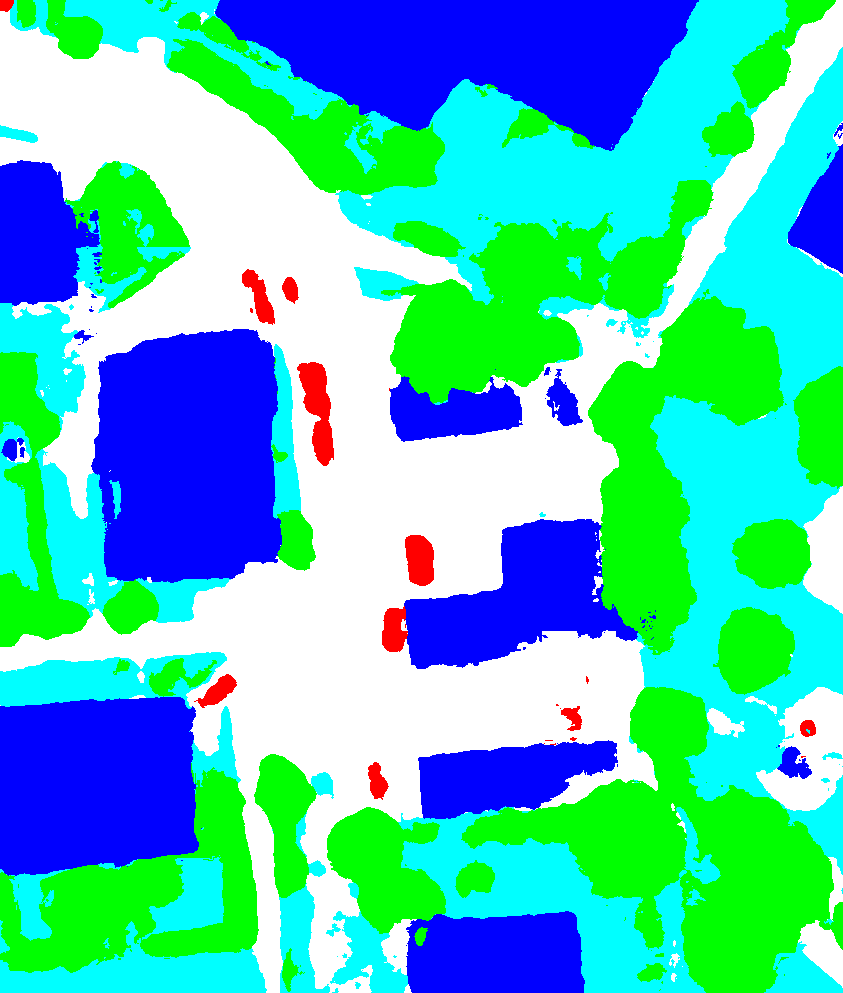}\\
\includegraphics[width=.17\linewidth]{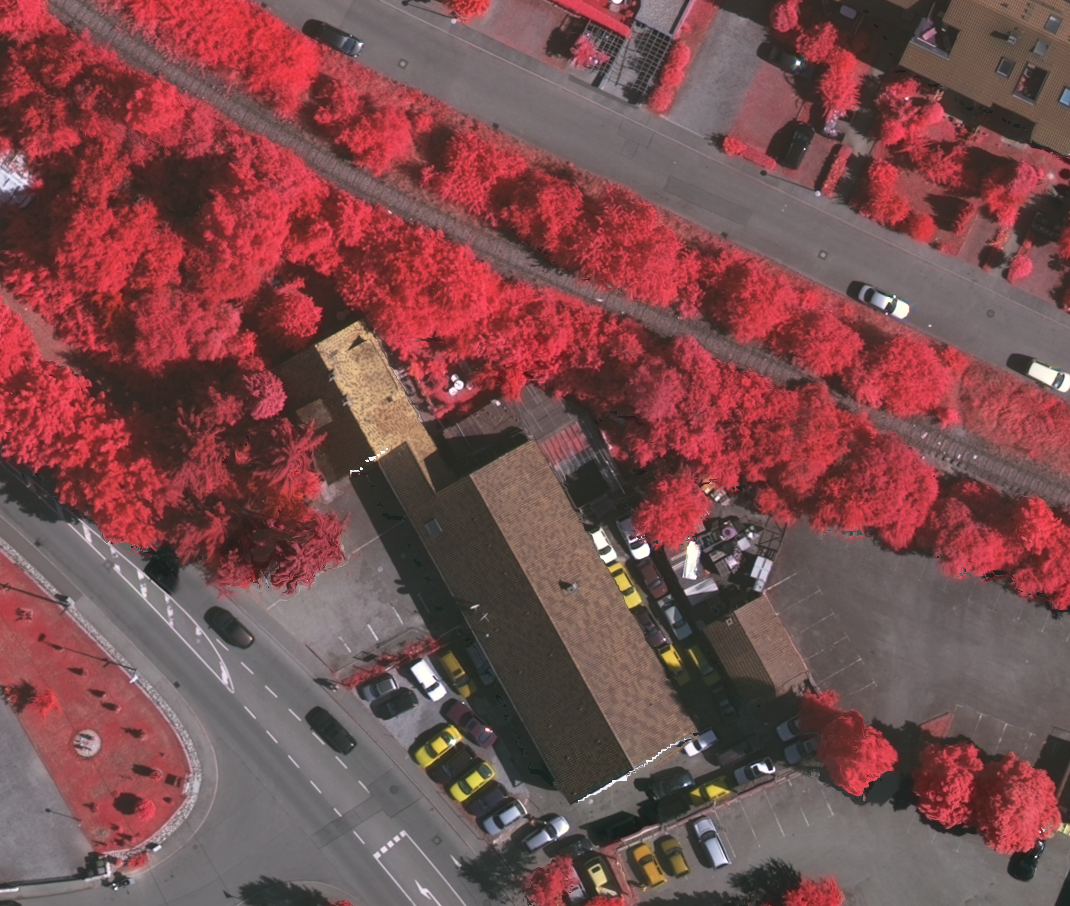} &
\includegraphics[width=.17\linewidth]{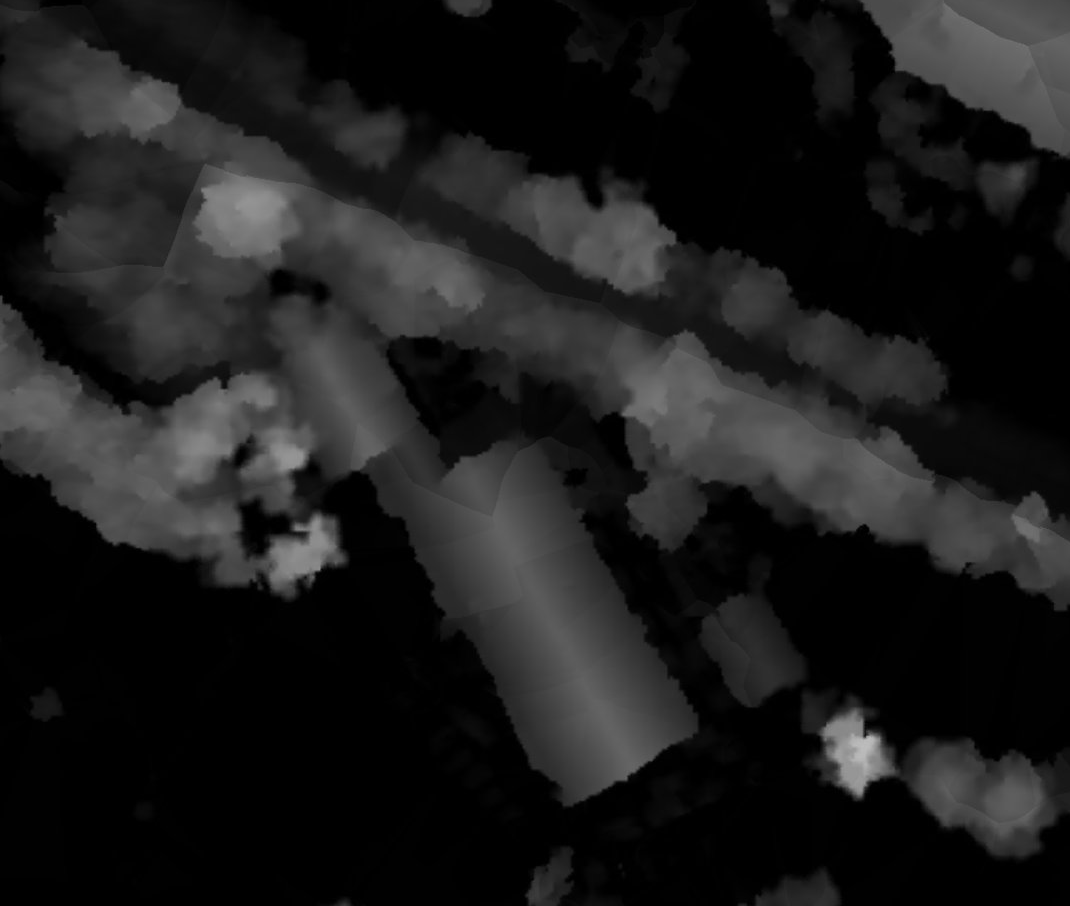} &
\includegraphics[width=.17\linewidth]{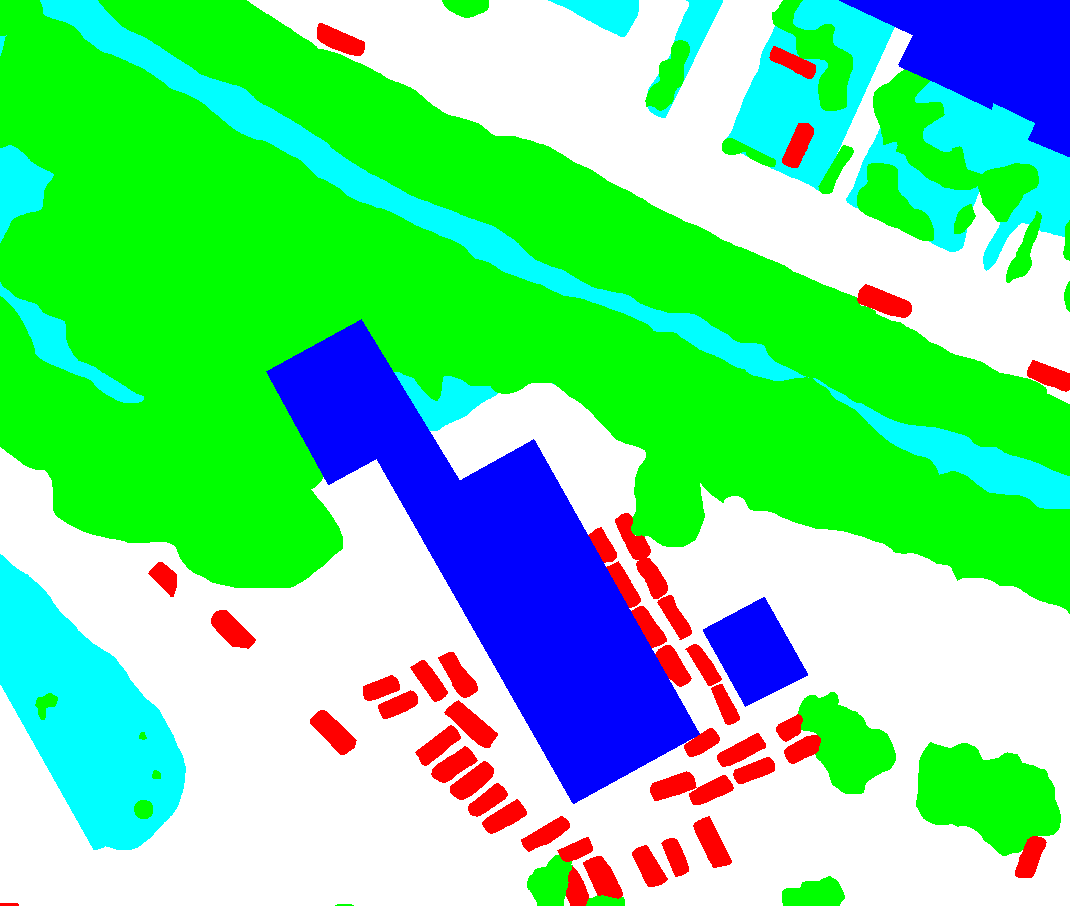} &
\includegraphics[width=.17\linewidth]{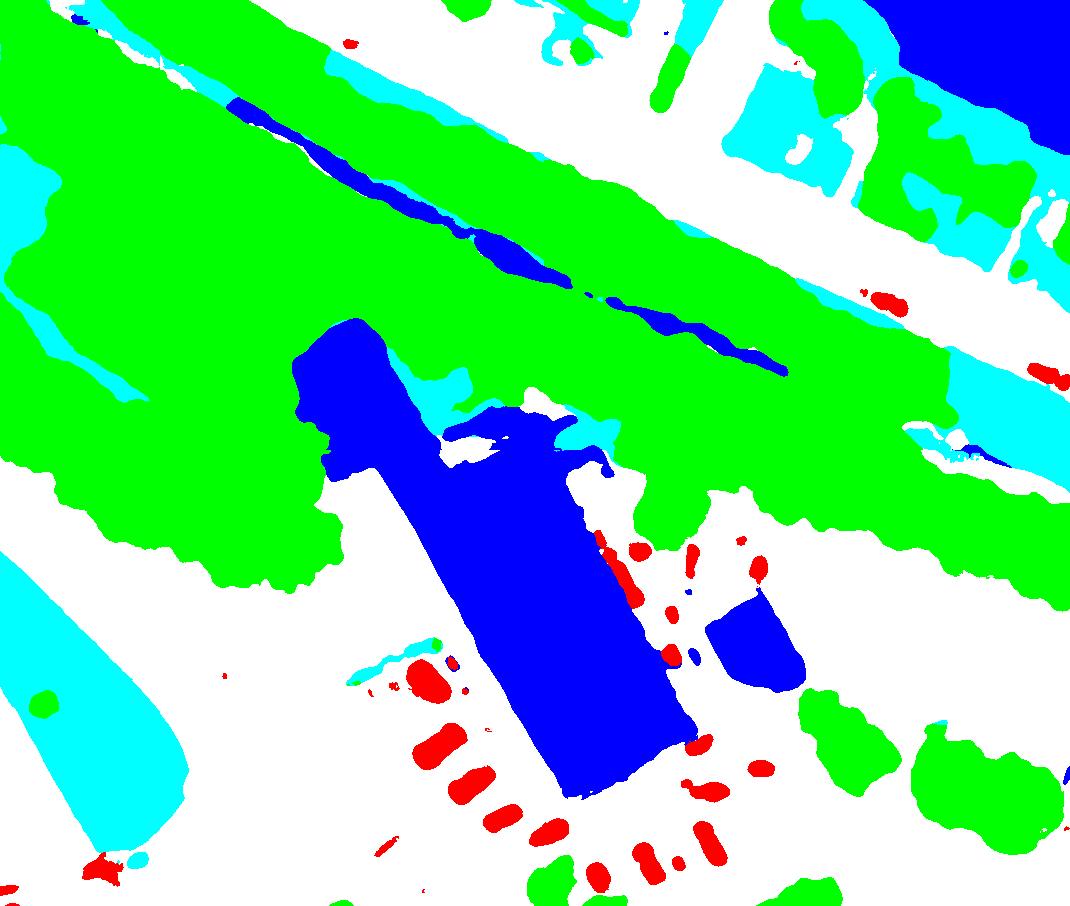} &
\includegraphics[width=.17\linewidth]{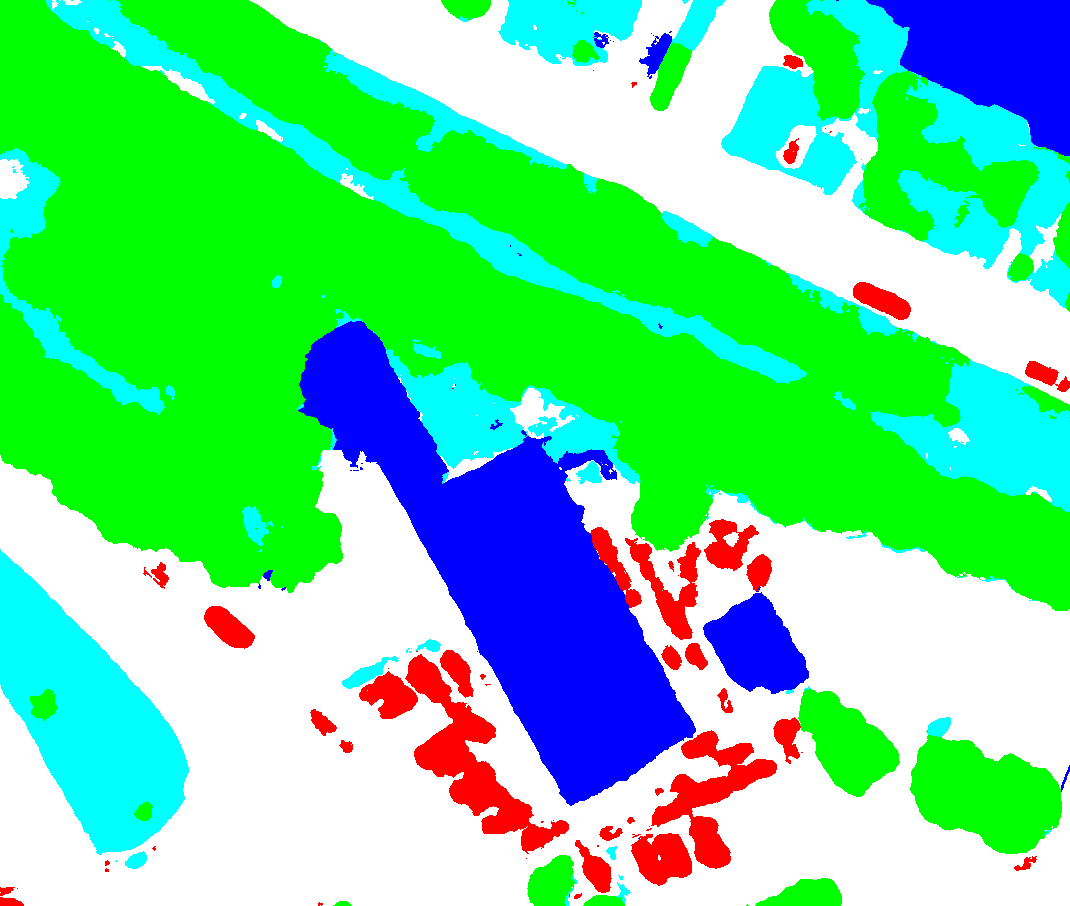}\\
\includegraphics[width=.17\linewidth]{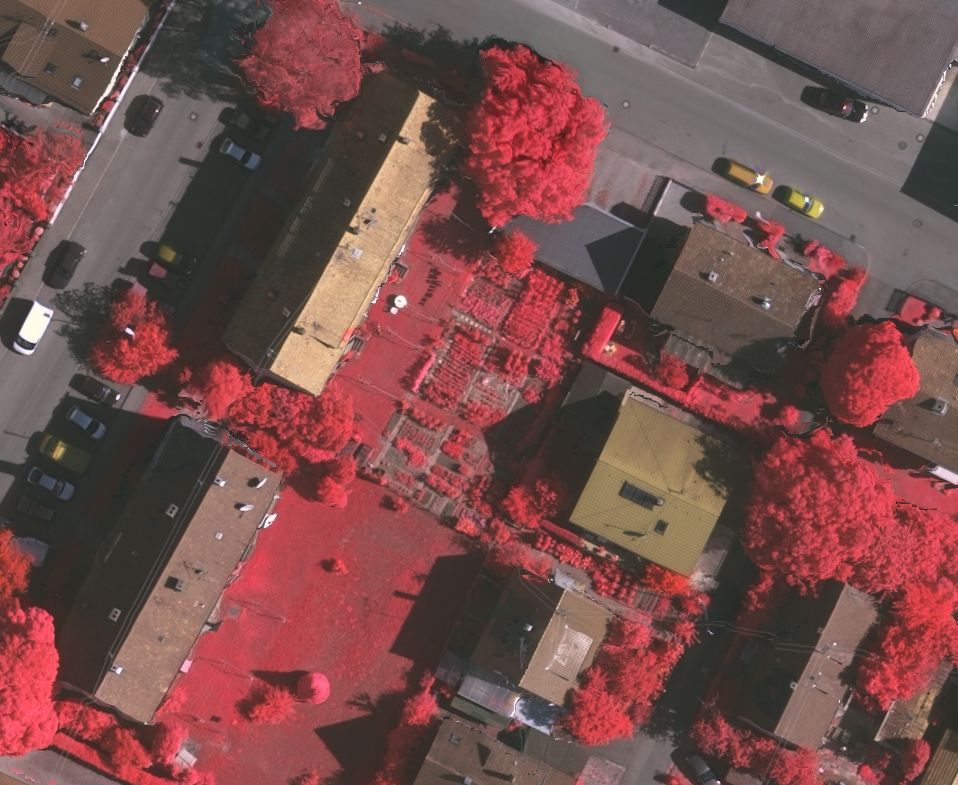} &
\includegraphics[width=.17\linewidth]{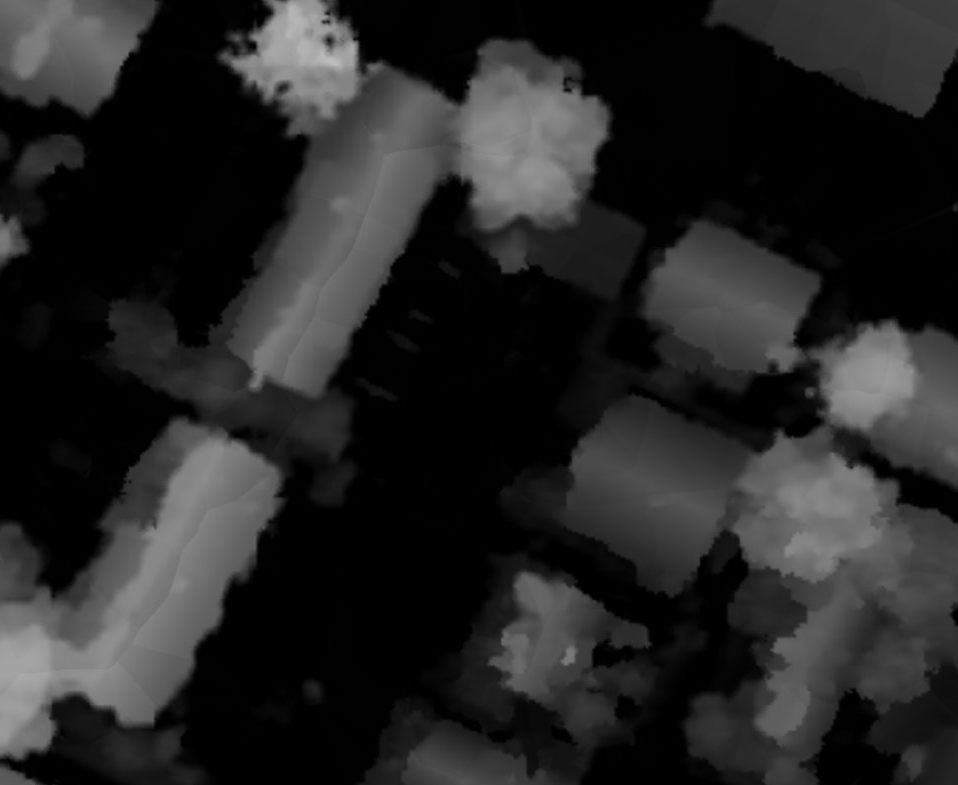} &
\includegraphics[width=.17\linewidth]{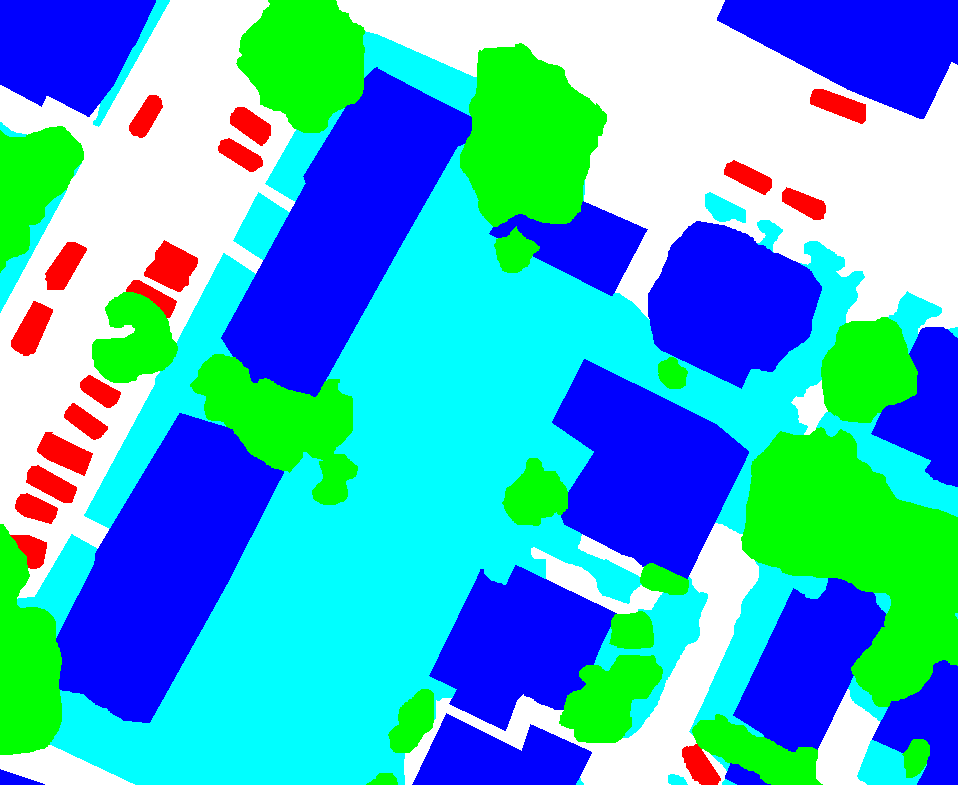} &
\includegraphics[width=.17\linewidth]{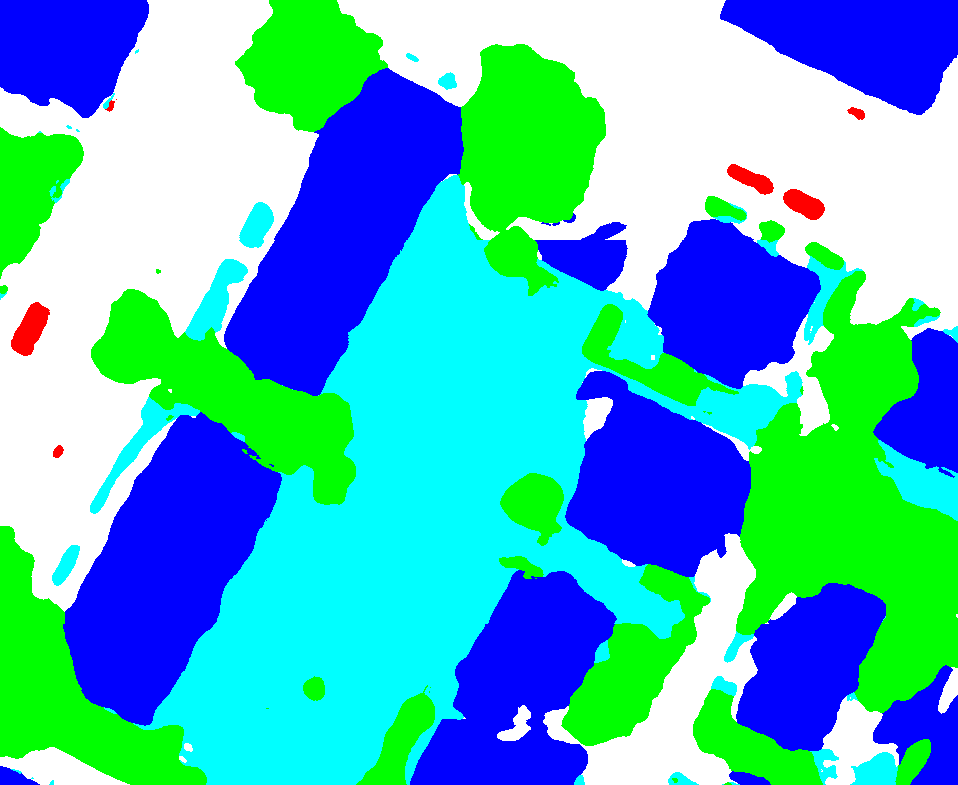} &
\includegraphics[width=.17\linewidth]{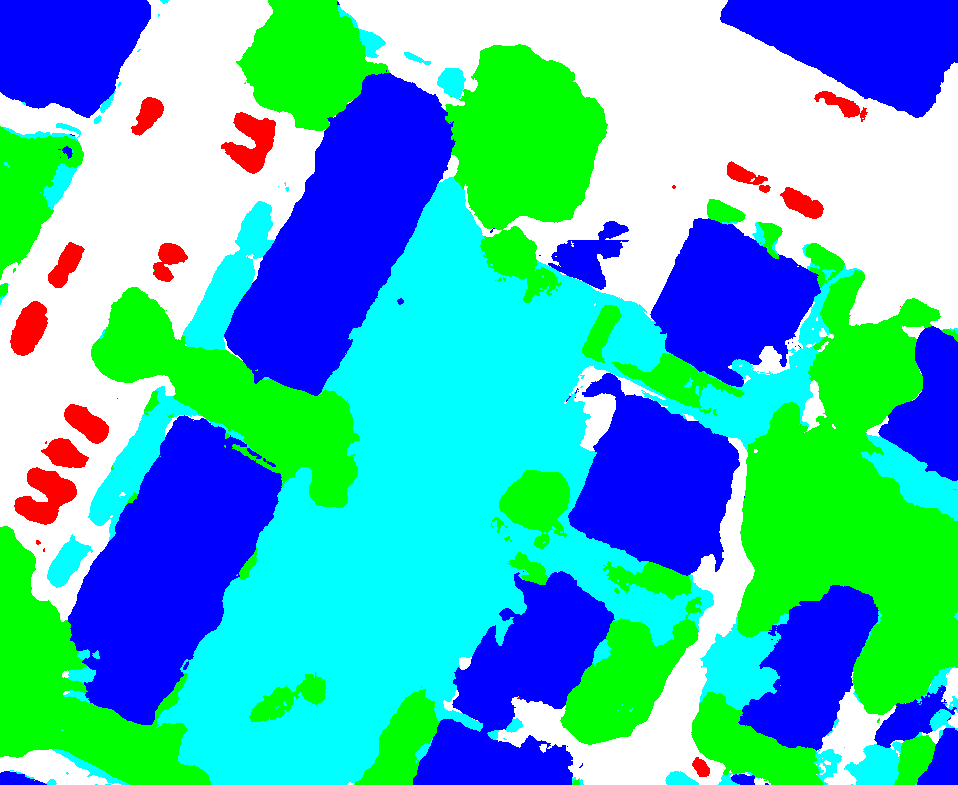}\\
\end{tabular}
\caption{Examples of classification maps obtained in the Vaihingen validation images with the RotEqNet and the standard CNN models.}
\label{fig:maps}
\end{figure}

Fig. \ref{fig:maps} shows a few examples of the obtained classification maps. We see how RotEqNet performs better on smaller objects, such as cars or the grass path in the second image, but generates less smooth edges. The latter is possibly due to different orientation for certain features being chosen in contiguous pixels.

\subsection{Zeebruges}

The results in Table~\ref{tab:zee} show the performance of the proposed method on the Zeebruges dataset compared to the last published results in~\citet{DFCA}. Although the authors of~\citet{DFCA} were not aiming at obtaining lightweight models, the two best results they report are obtained by CNNs containing of the order of $10^7$ parameters, while RotEqNet achieves comparable results with a mere $10^5$ parameters, two orders of magnitude less. In particular, out the models used by~\citet{DFCA}, the VGG/SVM model consists of a linear SVM classifier trained on features extracted by a VGG network with $2.5\cdot10^7$ parameters, while the AlexNet model is a pretrained CNN that has been fine tuned end-to-end on the benchmark's training set. It has around $6\cdot10^7$ parameters. A RotEqNet network with $1.4\cdot10^5$ parameters, with Nf=4, was enough to obtain better results than the VGG/SVM model, and one with $4.3\cdot10^5$ parameters, Nf=7, was enough to close the accuracy gap with the fine tuned AlexNet. These results highlight the advantage in terms of model size reduction that can be obtained by sparing the model from having to learn to be equivariant to rotations. In this dataset we see again that RotEqNet performs particularly well on the car and the building classes, both associated with strong edge features, while it lags behind with respect to both competing models in the tree class, which contains rather anisotropic features.

\begin{table}[h]
	\centering
	\caption{Results on Zeebruges. F1 scores per class and global average (AA) and overall accuracies (OA) and Cohen's Kappa. Best result per row is in dark gray, second in light gray.}
	\label{tab:zee}
	\begin{tabular}{c|ccc|cc}
		\hline
		 Model    	& \multicolumn{3}{c|}{RotEqNet} & VGG/SVM$^*$ & AlexNet$^*$ \\\hline\hline
		 \# parameters & $1.4\cdot10^5$ & $2.2\cdot10^5$& $4.3\cdot10^5$ & $2.5\cdot10^7$ & $6\cdot10^7$\\
		 \hline\hline

Impervious		&	74.75 & 74.98 & 77.41\cellcolor[gray]{0.8}	&	67.66	&	\textbf{79.10}\cellcolor[gray]{0.6}	\\
Water		&	98.46 & 98.56\cellcolor[gray]{0.8}	& \textbf{98.69}\cellcolor[gray]{0.6} &	96.50	&	98.20	\\
Clutter		&	31.19 & 36.27 & 48.99\cellcolor[gray]{0.8}	&	45.60	&	\textbf{63.40}\cellcolor[gray]{0.6}	\\
Low Vegetation	&	76.58 & 77.89& \textbf{78.73}\cellcolor[gray]{0.6} &	68.38	&	78.00\cellcolor[gray]{0.8}	\\
Building		&	69.26 &	75.11& \textbf{79.07}\cellcolor[gray]{0.6} &	72.70	&	75.60\cellcolor[gray]{0.8}	\\
Tree		&	59.35 &	68.95& 71.30 &	78.77\cellcolor[gray]{0.8}	&	\textbf{79.50}\cellcolor[gray]{0.6}	\\
Boat		&	39.13 &	43.59& 44.55	& \textbf{56.10}\cellcolor[gray]{0.6}	&	44.80\cellcolor[gray]{0.8}	\\
Car		&	56.26\cellcolor[gray]{0.8}	& \textbf{56.61}\cellcolor[gray]{0.6} & 52.54 &	33.90	&	50.80	\\\hline
		OA  & 79.2 & 80.8& 82.6\cellcolor[gray]{0.8} & 76.6 & \textbf{83.32}\cellcolor[gray]{0.6} \\ 
		AA &   69.2 & 73.2\cellcolor[gray]{0.8} & 75.3\cellcolor[gray]{0.6} & - & -   \\ 
		Kappa  & 0.73 & 0.75&  0.77\cellcolor[gray]{0.8} & 0.70 & \textbf{0.78}\cellcolor[gray]{0.6} \\\hline
	\multicolumn{6}{l}{$^*$ = from \citet{DFCA}}
		\end{tabular}
\end{table}

\subsection{Computational time}

On the one hand, due to the additional burden of requiring to interpolate the rotated filters and the linear dependency between the number of orientations $R$ and the number of convolutions to compute, RotEqNet can potentially increase the computational time required with respect to a standard CNN. On the other hand, the reduction in the number of feature maps, which is independent of $R$, can compensate for this if $R$ is small enough. As we can see in Table~\ref{tab:time}, the RotEqNet model tested on the Vaihingen validation set and trained with $R=16$ outperforms the standard CNN in terms of speed up to $R=64$. Note that all tests are performed on a single CPU to make the results more comparable.

\begin{table}[h]
	\centering
	\caption{Computational time of the forward pass in a single CPU and accuracy on the validation set of the Vaihingen dataset. The RotEqNet models are tested with different values of the number of orientations, $R$.}
	\label{tab:time}
	\begin{tabular}{c|ccccc|c}
		\hline
		 Model    	& \multicolumn{5}{c|}{RotEqNet} & CNN  \\\hline\hline
		 $R$ & $8$ & $16$& $32$ & $64$ & $128$ & -\\
		 \hline\hline

OA		&	86.90 & 87.89 & 87.81	&	87.71	&	87.65 & 87.47	\\
AA		&	80.69 & 84.34	& 85.18 &	85.33	&	85.51 & 78.18	\\
Kappa		&	0.82 & 0.84 & 0.84	&	0.83	&	0.83  & 0.83	\\
	\hline
		Time per tile (s)  & 1.4 & 1.7 & 2.3  & 3.1 & 5.0 & 4.2  \\\hline
		\end{tabular}
\end{table}

\section{Conclusion}

Deep learning models, and in particular convolutional neural networks, have shown their potential for remote sensing image analysis. By learning filters directly from data, they allow to learn and encode spatial information without engineering the feature space in a problem-dependent way. But if these models have potential, they still suffer from the need for an extensive (and comprehensive) training set, cumbersome hyperparameter tunning and considerable computing resources, both in terms of memory and operations. 
In this paper, we have explored the possibility of reducing such requirements by encoding one prior information about the images: the fact that their orientation, as well as that of objects it contains, is often arbitrary. Such prior can be exploited by making the CNN model rotation equivariant, i.e. by forcing the network to react in the same way each time it encountered the same semantic class, independently from the spatial orientation of the features. We achieved this behavior by applying rotating convolutions, where a canonical filter is applied at many orientations and the maximal activation is propagated through the CNN. The proposed RotEqnNet therefore has minimal memory and storage requirements, since it does not need to learn filters which respond to each specific orientation and thus generates less intermediate feature maps at runtime. Rotation equivariance is encoded within the model itself (similarly to how CNNs achieve translation equivariance) and propagating only maximal activations reduces the model size and runtime memory requirements while keeping most of the orientation information. 

We applied the proposed framework to two subdecimeter land cover semantic labeling benchmarks. The results show two main tendencies: on one hand, that explicitly encoding rotation equivariance in deep learning dense semantic labeling models allows for much smaller models, between one and two orders of magnitude compared to traditional CNNs. 
On the other hand, they also show that a CNN encoding equivariance in its structure -- rather than through data augmentation -- also provides robustness against varying amounts of training data, allowing to train efficiently and perform well in modern remote sensing tasks. This last point is of particular importance when considering that the amount of available labels can vary enormously in remote sensing depending on the mode of acquisition and the problem at hand. 

RotEqNet is not limited to semantic labeling tasks. Its logic can be applied to any deep model involving convolutions where a predefined behavior with respect to rotations is expected. As shown in~\citet{marcos2016rotation}, it can be applied to various applications requiring rotation invariance, equivariance or even covariance, which opens doors for the application of RotEqNet to tackle problems of detection (cars, airplanes, trees) or regression (super-resolution, biophysical parameters) when only limited  labeled instances are at hand.





\section*{Acknowledgements}
This work has been partly supported by the Swiss National Science Foundation (grant PZ00P2-136827, http://p3.snf.ch/project-136827). The authors would like to thank the Belgian Royal Military Academy for acquiring and providing the 2015 Contest data, and the ONERA -- The French Aerospace Lab -- for providing the corresponding ground-truth. They also would like to thank the ISPRS working group II/4 for providing the data of the 2D semantic labeling challenge over Vaihingen.

\bibliographystyle{elsarticle-harv}
\biboptions{authoryear}
\bibliography{biblio}

\end{document}